\documentclass{article}

\usepackage[preprint,nonatbib]{dms} % uncomment acknowledments and change Code and Data Availability for arXiv

\usepackage[utf8]{inputenc} % allow utf-8 input
\usepackage[T1]{fontenc}    % use 8-bit T1 fonts
\usepackage{hyperref}       % hyperlinks
\usepackage{url}            % simple URL typesetting
\usepackage{booktabs}       % professional-quality tables
\usepackage{amsfonts}       % blackboard math symbols
\usepackage{nicefrac}       % compact symbols for 1/2, etc.
\usepackage{microtype}      % microtypography
\usepackage{xcolor}         % colors
\usepackage{subcaption}     % for subfigures
\usepackage{graphicx}       % for includegraphics

\usepackage{pifont}% http://ctan.org/pkg/pifont (for \ding)
\newcommand{\cmark}{\ding{51}}%
\newcommand{\xmark}{\ding{55}}%
\newcommand{\bld}[1]{\textbf{#1}}

\title{DirectMultiStep: Direct Route Generation for Multistep Retrosynthesis}

\author{%
  Yu Shee\thanks{Equal contribution. Listing order is random.} \\
  Yale University\\
  \texttt{yu.shee@yale.edu} \\
  \And
  Anton Morgunov* \\
  Yale University\\
  \texttt{anton@ischemist.com} \\
  \And
  Haote Li \\
  Yale University\\
  \texttt{haote.li@yale.edu} \\
  \And
  Victor Batista \\
  Yale University\\
  \texttt{victor.batista@yale.edu} \\
}

\begin{document}

\maketitle

\begin{abstract}
  Traditional computer-aided synthesis planning (CASP) methods rely on iterative single-step predictions, leading to exponential search space growth that limits efficiency and scalability. We introduce a series of transformer-based models, that leverage a mixture of experts approach to directly generate multistep synthetic routes as a single string, conditionally predicting each transformation based on all preceding ones. Our DMS Explorer XL model, which requires only target compounds as input, outperforms state-of-the-art methods on the PaRoutes dataset with 1.9x and 3.1x improvements in Top-1 accuracy on the n$_1$ and n$_5$ test sets, respectively. Providing additional information, such as the desired number of steps and starting materials, enables both a reduction in model size and an increase in accuracy, highlighting the benefits of incorporating more constraints into the prediction process. The top-performing DMS-Flex (Duo) model scores 25-50\% higher on Top-1 and Top-10 accuracies for both n$_1$ and n$_5$ sets. Additionally, our models successfully predict routes for FDA-approved drugs not included in the training data, demonstrating strong generalization capabilities. While the limited diversity of the training set may affect performance on less common reaction types, our multistep-first approach presents a promising direction towards fully automated retrosynthetic planning. 
\end{abstract}

\section{Introduction}

Finding the most efficient route to a desired chemical compound is a daily challenge for synthetic organic chemists in both fundamental research and drug discovery. Route efficiency is determined by various factors, some of which can be objectively assessed, such as overall yield (not all chemical reactions have 100\% conversion rate) and enantiomeric excess (in case of chiral compounds), where higher values are always preferred. Other factors, such as atom efficiency (minimization of byproducts) and availability (cost) of starting materials, are more case-dependent. A used chemical reactant is considered waste unless it can be repurposed as a reactant in a different process. Similarly, the choice of starting materials depends on factors such as budget, logistics, and the availability of specific equipment. It's worth noting that many commercially available compounds can be synthesized from other commercially available compounds, adding another layer of complexity to the decision-making process.

Algorithmic frameworks for identifying synthetic routes were envisioned by Vleduts~\cite{vleduts1963concerning} and further formalized by Elias James Corey (subject of the 1990 Nobel Prize in Chemistry) into what is now known as retrosynthetic analysis. This framework begins with identifying atoms that would serve as reaction centers. Disconnecting bonds between these centers results in the formation of hypothetical fragments (called synthons) from which a precursor molecule can be created. This mapping from synthons to actual molecules is one-to-many because there is usually more than one functional group that could participate in any given type of reaction. Importantly, meticulous application of Corey's framework (i.e., systematically breaking small subsets of bonds) will eventually lead to commercially available starting materials. The algorithmic nature of this process allowed Corey to envision automating these rules to create Computer-Aided Synthesis Planning (CASP) as early as 1969~\cite{corey_computer-assisted_1969}.

Recent advancements in data science and machine learning (ML) methods have led to a surge of interest in developing CASP methods~\cite{de_almeida_synthetic_2019, struble_current_2020}. The vast majority of existing methods~\cite{guo_bayesian_2020, lee_retcl_2021, segler_neural-symbolic_2017, coley_computer-assisted_2017, ishida_prediction_2019, fortunato_data_2020, dai_retrosynthesis_2020, chen_deep_2021, seidl_improving_2022, yan_retroxpert_2020, shi_graph_2021, somnath_learning_2021, wang_retroprime_2021, wang_retrosynthesis_2023, zhong_retrosynthesis_2023, liu_retrosynthetic_2017, karpov_transformer_2019, chen_learning_2019, lee_molecular_2019, lin_automatic_2020, zheng_predicting_2020, tetko_state---art_2020, seo_gta_2021, mao_molecular_2021, sacha_molecule_2021, mann_retrosynthesis_2021, ucak_substructure-based_2021, kim_valid_2021, irwin_chemformer_2022, zhong_root-aligned_2022, ucak_retrosynthetic_2022, shee2024site} are designed to automate single-step retrosynthetic (SSR) analysis, i.e. predicting a list of compounds from which a target product could be made in one step. A full multistep route can be determined by iteratively applying SSR methods to the identified precursors until a termination condition is satisfied (e.g., identifying reactants in the database of commercially available compounds). Notably, because each SSR method call creates a list of candidates, iterative application of these methods generates an exponentially growing search space. Graph traversal algorithms such as Monte Carlo Tree Search (MCTS), Depth-First Proof Number (DFPN) search, and A-star (Retro*) have been applied to efficiently traverse this exponential search space~\cite{segler_planning_2018,kishimoto_depth-first_2019,chen_retro_2020}. More sophisticated search designs include utilization of hyper-graph~\cite{schwaller2020predicting}, graph-based neural networks~\cite{xie2022retrograph}, reinforcement learning~\cite{yu2022grasp}, or incorporation of previous reactions as context for single-step predictions~\cite{liu2023fusionretro}.

Evaluation of performance of these SSR methods on full route prediction has been limited by the scarcity of open-source datasets containing valid multistep routes. The creation of PaRoutes~\cite{genheden_paroutes_2022}, an open-source dataset containing 450k multistep routes (163k excluding duplicates and permutations, along with two test sets of 10k routes each) extracted from the United States Patent and Trademark Office (USPTO), marked a major development. Notably, even state-of-the-art SSR models~\cite{segler_neural-symbolic_2017, thakkar_datasets_2020} combined with advanced search algorithms (MCTS, DFPN, Retro*) correctly identify multistep routes (Top-1 accuracy) for only 17\% and 10\% of the target compounds in n$_1$ and n$_5$ test sets, respectively. Such performance could be rationalized by recognizing that retrosynthesis is inherently a multistep problem: the optimal choice of reaction to make a compound depends on subsequent steps in the synthesis. For example, a common pattern in multistep routes includes: (1) protection of certain functional groups, (2) the desired transformation of unprotected groups, and (3) deprotection of the protected groups. An SSR method applied to the reactant of reaction (2) may output numerous candidate precursors; however, knowing that protective groups are removed in subsequent steps dramatically changes the probability distribution over those precursors.

In this work, we propose a novel approach for direct prediction of multistep routes, bypassing the need for single-step models and sophisticated exponential graph traversal algorithms. We begin with a brief overview of the methodology and evaluation metrics. We proceed by demonstrating and discussing performance of DMS models on diverse n$_1$ and n$_5$ evaluation sets (10 000 routes each). We demonstrate further generalizability by successfully predicting experimental synthetic routes for FDA-approved drugs. We conclude by discussing the limitations of our approach. A more elaborate description of model architectures and rationale for certain design choices is given at the end of the paper.

\section{Methodology Overview}
\subsection{Definitions}

\begin{figure}[htbp]
  \centering
  \includegraphics[width=\textwidth]{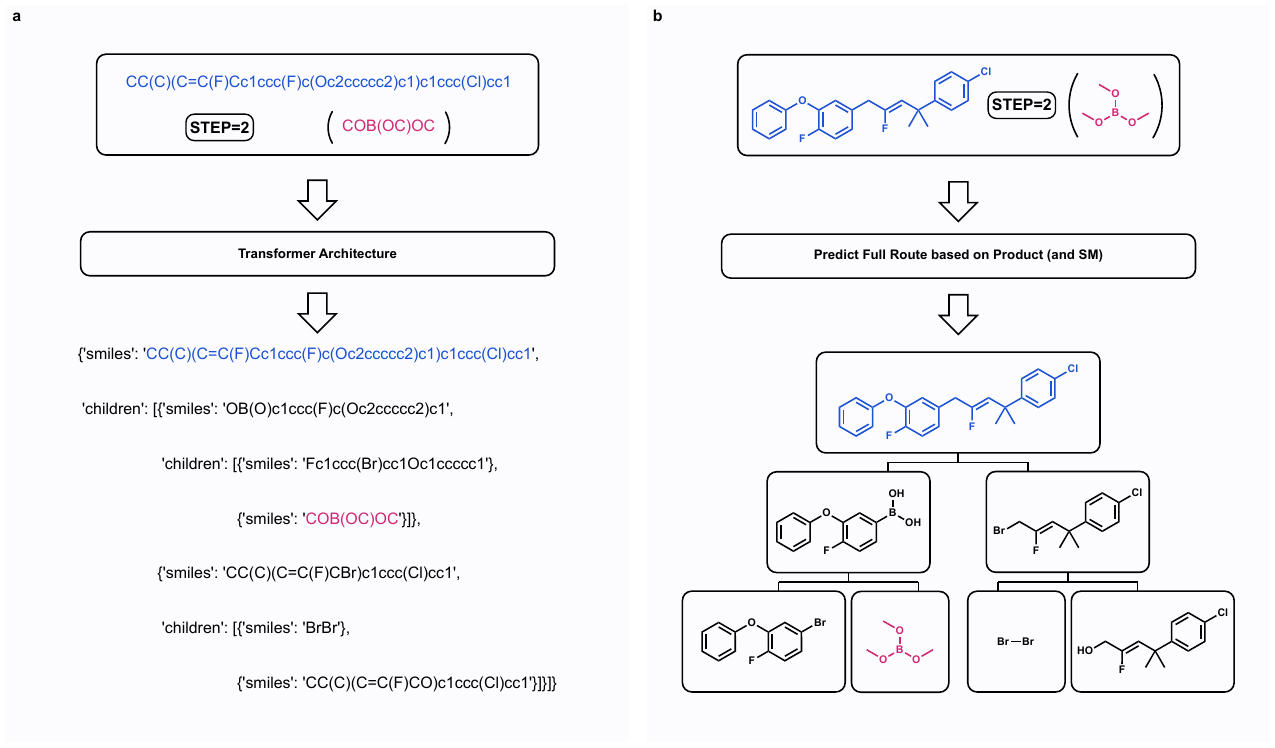}  
    \begin{subfigure}{0pt} 
        \phantomcaption
        \label{fig:method_overview_code}
    \end{subfigure}%
    
    \begin{subfigure}{0pt} % Empty subfigure for phantom label
        \phantomcaption
        \label{fig:method_overview_graphic}
    \end{subfigure}
  \caption{The workflow of DirectMultiStep. (a) The SMILES representation of the target compound (blue), starting material (red, optional), and the number of steps (optionally) are tokenized, concatenated, and fed into our transformer model. The model predicts a string representation of the multistep synthesis tree. Spaces are added for clarity, and indentations indicate the levels in the synthesis route (tree). (b) Molecular structures corresponding to the target compound (blue), starting material (red, optional), and the predicted synthesis tree with structures of all molecules.}
  \label{fig:method_overview}
\end{figure}

\textbf{Target compound} (blue in Fig.~\ref{fig:method_overview}) is the final (desired) product of the multistep synthesis tree. \textbf{Starting material, SM} (red in Fig.~\ref{fig:method_overview}) is a compound for which no further precursors need to be identified, i.e., a leaf of the synthesis tree. The \textbf{number of steps} is the largest number of reactions from SM to the target compound (tree height).

\subsection{Route Representation}
Routes are represented as recursive dictionaries (Fig.~\ref{fig:method_overview_code}) containing the SMILES representation of the molecule and a list of other dictionaries, containing either starting materials or trees leading to precursors of the current node. Removing space and newline characters creates a string representation. Our models predict that string by taking SMILES of the target compound, (optionally) starting material, and (optionally) number of steps as input.

\begin{figure}[htbp]
  \centering
  \includegraphics[width=\textwidth]{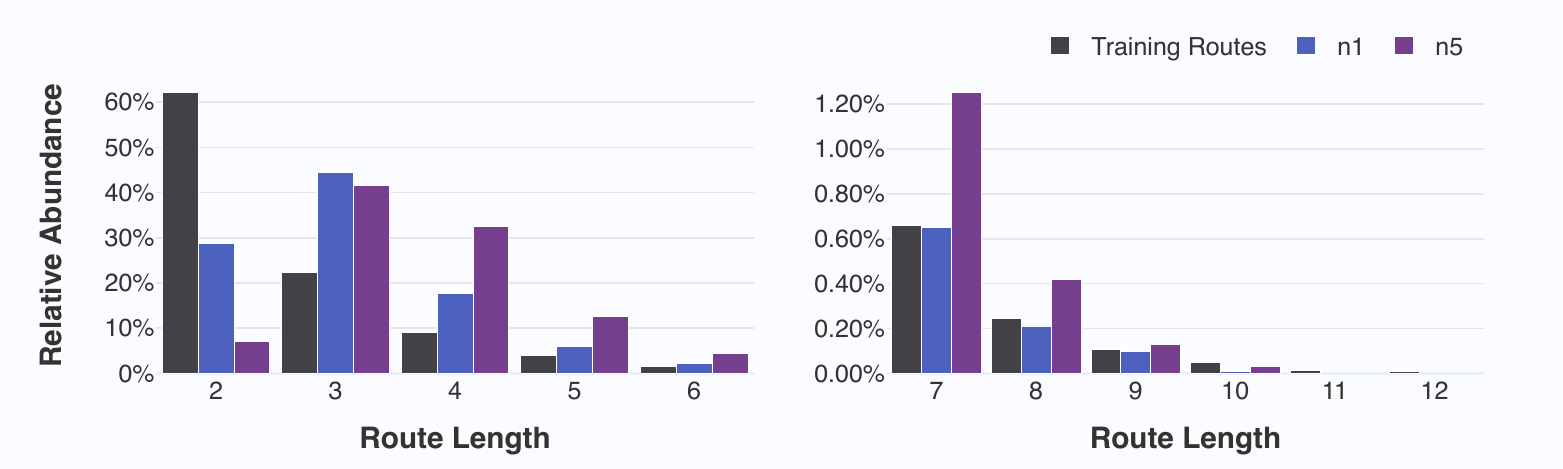}  
  \caption{Distribution of the relative frequencies of route lengths (in terms of number of steps) in the training dataset before augmentation with permutations (163 689 routes, black), n$_1$ test set (10 000 routes, blue), and n$_5$ test set (10 000 routes, purple). Distribution is split into routes shorter (left subplot) and longer than 6 steps (right subplot).}
  \label{fig:length_distribution}
\end{figure}

\subsection{Route Generation and Post-Processing}

The structure of target compound, starting material (optionally), and number of steps (optionally) is passed to the transformer encoder. A beam search with a width of 50 is employed to generate 50 candidate routes by sampling next tokens from the decoder. Within beam search, during the prediction of each token the cumulative log likelihoods are normalized by the sequence length at the time the new token is sampled. This adjustment ensures that shorter routes with simpler compounds are not disproportionately favored while maintaining a balance between route length and overall sequence probability. After beam search finds end-of-sequence tokens for all 50 candidates, predicted routes are checked for validity of all SMILES strings (which are canonicalized), the presence of starting materials in the stock set, and absence of repetitions (which can be present as permutations of other routes). Remaining routes are compared against the true (experimental) route. The reproduction of experimental routes is presented as Top-$K$ accuracy, i.e. presence of correct route ranked $K$ or lower.

\section{Results and Discussion}

\subsection{Overview}
In this work, we present DMS Explorer XL, a mixture-of-experts based model (50M parameters), that predicts full multistep retrosynthetic route given the structure of the target compound. Our model shows comparable search performance on ChemBL-5000\cite{genheden2021clustering} and significantly improves Top-1 accuracy on n$_1$ and n$_5$ evaluation datasets from PaRoutes\cite{genheden_paroutes_2022}. We find that if one is willing to provide extra information to the encoder, such as desired route length and (or) starting material, the model size can be reduced and simultaneously both Top-1 and Top-10 accuracy can be increased significantly. We present a family of such models (model cards are in Sec.~\ref{sec:model-cards}) and discuss their use cases.

\subsection{Search Performance}
First, we report the solved rate on ChemBL-5000~\cite{genheden2021clustering} (Table \ref{table:search_performance_chembl}). Our baseline DMS Explorer XL (50M) shows comparable performance to that of a tree search on predictions from a single-step model (AiZynthFinder~\cite{genheden2021clustering}). A slightly smaller DMS Wide (40M) model that was trained with desired route length as an extra parameter to the encoder shows significantly better performance. Despite 7x more calls to the model (with different route length parameters), the average run time per target is only 3x higher. Notably, a 4x smaller DMS Flash (10M) model trained with both route length and structure of the starting material, can be evaluated on new targets even without the structure of the starting material with a minor (3-5\%) decrease in the solved rate, but at the same cost as the DMS Explorer XL model.

\begin{table}[htbp]
	\caption{Search performance for DirectMultiStep models on ChemBL-5000~\cite{genheden2021clustering}}
	\centering
	\begin{tabular}{llcc}
		\toprule
		Method$^{a}$                                & Stock Set$^{b}$      & Solved Rate & Run Time (s)$^c$ \\
        \midrule
		DMS Explorer XL                             & Buyables             & 68.66\%     & 26.9             \\
		AiZynthFinder~\cite{genheden2021clustering} & In-House + Enamine   & 75.24\%     & NA               \\
		DMS Explorer XL                             & eMolecules Screening & 75.58\%     & 26.9             \\
		DMS Flash$^{d,e}$                    & Buyables             & 86.66\%     & 23.1             \\
		DMS Flash$^{d,e}$                    & eMolecules Screening & 88.54\%     & 23.1             \\
		DMS Wide$^d$                         & Buyables             & 91.20\%     & 75.4             \\
		DMS Wide$^d$                         & eMolecules Screening & 91.94\%     & 75.4             \\
		\bottomrule
	\end{tabular}
	\label{table:search_performance_chembl}
	\begin{minipage}{\textwidth}
		\raggedright
		\footnotesize{$^a$ All DMS models are run with a beam size of 50 on a single NVIDIA A100 GPU with half-precision floating point inference (FP16).} 
		\\\footnotesize{$^b$ eMolecules Screening is from Chen et al.~\cite{chen_retro_2020} and contains 23.1M screening compounds from eMolecules as of 2019 (2019-11-01). Buyables is from Roh et al.~\cite{roh2025higher} and includes 0.329M buyable building blocks from eMolecules, Sigma-Aldrich, Mcule, ChemBridge Hit2Lead, and WuXi LabNetwork. In-House + Enamine is from Genheden et al.~\cite{genheden2021clustering} and is not publicly available.}  
		\\\footnotesize{$^c$ Averages over all targets. NA indicates that the run time is not available in the original publication.}  
		\\\footnotesize{$^d$ Uses step counts from 2 to 8 (total of 7 DMS model runs).}
		\\\footnotesize{$^e$ SMs not provided for DMS-Flash.}
		
	\end{minipage}
\end{table}

Next, we report the solved rate on n$_1$ and n$_5$ evaluation sets (each contains 10 000 routes) in Table~\ref{table:search_performance_paroutes}. Following the procedure of PaRoutes~\cite{genheden_paroutes_2022} the starting material stock set is taken to be the set of all leaves in n$_1$ or n$_5$. DMS Explorer XL performs noticeably worse than MCTS or Retro*, but scores better than DFPN on n$_5$. The number and the distribution of solved targets before enforcing the stock set is shown in Tables S1-S2 and Figures S1-S2. Other DMS models that accept desired route length (Wide) and also the structure of the starting material (Flash) solve more targets, but still underperform Retro*.

Unfortunately, the training set of PaRoutes contains routes that include 47 of the targets from the commonly used USPTO-190 evaluation subset, which makes direct comparison with existing models unfair. For completeness, we report solved rate on all 190 and 143 (excluding the 47) targets in Table S3.
\begin{table}[htbp]
	\caption{Search performance for DirectMultiStep models on n$_1$ and n$_5$ evaluation sets}
	\centering
	\begin{tabular}{lcccc}
		\toprule
		Method$^{a}$                         & Evaluation Set & Solved Targets & First Solution$^b$ & Search Time (s)$^b$ \\
		\midrule
		DMS-Explorer-XL                      & n$_1$          & 8008           & NA                & 14.7                \\
		DMS-Wide                             & n$_1$          & 8089           & NA                & 6.0                 \\
		DFPN~\cite{genheden_paroutes_2022}   & n$_1$          & 8475           & 43.0               & 347.3               \\
		DMS-Flash                            & n$_1$          & 8814           & NA                & 2.0                 \\
		MCTS~\cite{genheden_paroutes_2022}   & n$_1$          & 9714           & 8.6                & 303.3               \\
		Retro*~\cite{genheden_paroutes_2022} & n$_1$          & 9726           & 7.0                & 300.7               \\
		\midrule
		DFPN~\cite{genheden_paroutes_2022}   & n$_5$          & 7382           & 53.2               & 297.9               \\
		DMS-Explorer-XL                      & n$_5$          & 7904           & NA                & 16.3                \\
		DMS-Wide                             & n$_5$          & 7950           & NA                & 7.4                 \\
		DMS-Flash                            & n$_5$          & 8646           & NA                & 2.5                 \\
		MCTS~\cite{genheden_paroutes_2022}   & n$_5$          & 9676           & 11.7               & 365.7               \\
		Retro*~\cite{genheden_paroutes_2022} & n$_5$          & 9703           & 10.5               & 349.2               \\
		\bottomrule
	\end{tabular}
	\label{table:search_performance_paroutes}
	\begin{minipage}{\textwidth}
		\raggedright
		\footnotesize{$^a$ All DMS models are run with a beam size of 50 on a single NVIDIA A100 GPU with half-precision floating point inference (FP16).}
		\\\footnotesize{$^b$ Averages over all targets.} 		
	\end{minipage}
\end{table}

\subsection{Top-K Accuracy (Route Quality)}
An attentive reader might notice the absence of a measure of a feasibility of proposed transformations during the discussion of solved rate above. One might argue that the number of solved targets only matters if the proposed transformations to the set of commercially available materials can be performed experimentally. In the absence of a cheap \textit{ab initio} tool to predict experimental feasibility of the proposed transformation, following Genheden et al.~\cite{genheden_paroutes_2022} and Liu et al.~\cite{liu2023fusionretro}, we utilize Top-K accuracy on reproducibility of experimentally verified routes as the proxy for route quality.

\begin{table}[htbp]
	\caption{Top-$K$ accuracy on route test set-n$_1$ (10 000 routes).}
	\label{tab:n1_comparison}
	\centering
	\begin{tabular}{llcccllllll}
		\toprule
		Method      &      &        &         &        & Top-1      &            &            &            & Top-5      & Top-10     \\
		\midrule
		MCTS$^a$    &      &        &         &        & 0.17       &            &            &            & 0.46       & 0.49       \\
		Retro*$^a$  &      &        &         &        & 0.15       &            &            &            & 0.41       & 0.45       \\
		DFPN$^a$    &      &        &         &        & 0.11       &            &            &            & 0.17       & 0.17       \\
		\midrule
		DMS         &      &        & SM      & SM     &            &            &            &            &            &            \\
		Models      & Size & Steps  & (Train) & (Gen.) & Top-1      & Top-2      & Top-3      & Top-4      & Top-5      & Top-10     \\
		\midrule 
            Explorer-XL    & 50M  & \xmark & \xmark  & \xmark & 0.32       & 0.40       & 0.44       & 0.45       & 0.46       & 0.48       \\
		Deep        & 41M  & \cmark & \xmark  & \xmark & 0.36       & 0.44       & 0.47       & 0.49       & 0.50       & 0.52       \\
		Explorer    & 19M  & \xmark & \cmark  & \cmark & 0.36       & 0.45       & 0.48       & 0.51       & 0.52       & 0.55       \\
            Explorer    & 19M  & \xmark & \cmark  & \xmark & 0.27       & 0.33       & 0.36       & 0.37       & 0.39       & 0.40       \\
            Flash       & 10M  & \cmark & \cmark  & \cmark & 0.41       & 0.50       & 0.54       & 0.56       & 0.58       & 0.60       \\
            Flash       & 10M  & \cmark & \cmark  & \xmark & 0.32       & 0.39       & 0.42       & 0.44       & 0.45       & 0.47       \\
		Wide        & 38M  & \cmark & \xmark  & \xmark & 0.38       & 0.46       & 0.49       & 0.50       & 0.51       & 0.53       \\
		Flex (Mono) & 19M  & \cmark & \cmark  & \cmark & 0.32       & 0.40       & 0.43       & 0.44       & 0.45       & 0.48       \\
		Flex (Duo)  & 19M  & \cmark & \cmark  & \cmark & \bld{0.43} & \bld{0.52} & \bld{0.55} & \bld{0.57} & \bld{0.58} & \bld{0.61} \\
		Flex (Duo)  & 19M  & \cmark & \cmark  & \xmark & 0.34       & 0.41       & 0.44       & 0.45       & 0.46       & 0.47       \\
		\bottomrule
	\end{tabular}
	\\
	\footnotesize{$^a$These data are collected from the 2.0 version of PaRoutes in their GitHub repository~\cite{genheden_paroutes_2022}.}
\end{table}

\begin{table}[htbp]
	\caption{Top-$K$ accuracy on route test set-n$_5$ (10 000 routes).}
	\label{tab:n5_comparison}
	\centering
	\begin{tabular}{llcccllllll}
		\toprule
		Method      &      &        &         &        & Top-1      &            &            &            & Top-5      & Top-10     \\
		\midrule
		MCTS$^a$    &      &        &         &        & 0.10       &            &            &            & 0.28       & 0.33       \\
		Retro*$^a$  &      &        &         &        & 0.10       &            &            &            & 0.30       & 0.36       \\
		DFPN$^a$    &      &        &         &        & 0.05       &            &            &            & 0.07       & 0.07       \\
		\midrule
		DMS         &      &        & SM      & SM     &            &            &            &            &            &            \\
		Models      & Size & Steps  & (Train) & (Gen.) & Top-1      & Top-2      & Top-3      & Top-4      & Top-5      & Top-10     \\
		\midrule 
        Explorer-XL    & 50M  & \xmark & \xmark  & \xmark & 0.31       & 0.38       & 0.41       & 0.42       & 0.43       & 0.45       \\
		Deep        & 41M  & \cmark & \xmark  & \xmark & 0.33       & 0.40       & 0.43       & 0.45       & 0.46       & 0.48       \\
		Explorer    & 19M  & \xmark & \cmark  & \cmark & 0.33       & 0.40       & 0.43       & 0.45       & 0.46       & 0.49       \\
        Explorer    & 19M  & \xmark & \cmark  & \xmark & 0.25       & 0.31       & 0.34       & 0.35       & 0.36       & 0.37       \\
        Flash       & 10M  & \cmark & \cmark  & \cmark & 0.37       & 0.44       & 0.47       & 0.49       & 0.51       & 0.53       \\
        Flash       & 10M  & \cmark & \cmark  & \xmark & 0.30       & 0.36       & 0.39       & 0.40       & 0.41       & 0.43       \\
		Wide        & 38M  & \cmark & \xmark  & \xmark & 0.35       & 0.42       & 0.44       & 0.46       & 0.47       & 0.49       \\
		Flex (Mono) & 19M  & \cmark & \cmark  & \cmark & 0.28       & 0.34       & 0.36       & 0.38       & 0.39       & 0.41       \\
		Flex (Duo)  & 19M  & \cmark & \cmark  & \cmark & \bld{0.39} & \bld{0.46} & \bld{0.49} & \bld{0.51} & \bld{0.52} & \bld{0.54} \\
		Flex (Duo)  & 19M  & \cmark & \cmark  & \xmark & 0.32       & 0.38       & 0.40       & 0.42       & 0.42       & 0.44       \\
		\bottomrule
	\end{tabular}
	\\
	\footnotesize{$^a$These data are collected from the 2.0 version of PaRoutes in their GitHub repository~\cite{genheden_paroutes_2022}.}
\end{table}

We report Top-$K$ accuracy on n$_1$ and n$_5$ sets, each containing 10 000 routes. Notably, both sets were constructed to maximize internal diversity (see the PaRoutes paper~\cite{genheden_paroutes_2022} for the procedure), and as seen in Fig. \ref{fig:length_distribution}, differ significantly from the training set in route length distribution. The DMS Explorer XL model (only target required) shows significant improvement on Top-1 accuracy for both n$_1$ (1.8x, Table \ref{tab:n1_comparison}) and n$_5$ (3.1x, Table \ref{tab:n5_comparison}) sets, while matching (n$_1$) or improving (n$_5$) performance on Top-10 accuracy.

We attribute this improvement to the multistep-first nature of our approach. To understand the distinction from existing approaches, it helps to notice that iterative application of single-step calls results in a breadth-first search-like exploration of a node: all reactants leading to a compound are generated at the same time. Our approach, however, is a depth-first search-like: the whole left sub-tree of target compound in Fig. \ref{fig:method_overview} is predicted before the immediate right child. This provides an extra context for the prediction of the second precursor to the target compound (or any other right child), which reduces the search space and finds more relevant compounds faster.

The superior performance of DMS models does not incur additional computational cost. A search with a beam size of 50 takes 1-40 seconds (depending on the complexity of the target and model size) on a single NVIDIA A100 GPU, which is comparable to the time required to find the first successful route using previous methods (7-50 seconds)~\cite{genheden_paroutes_2022}.

\subsubsection{Input information and model size trade-off}
A unifying pattern of results in Tables~\ref{tab:n1_comparison}-\ref{tab:n5_comparison} is that if one is willing to provide more information to the model, model size can be reduced and performance can be improved at the same time. For example, providing the desired route length reduces model size by 20\% and improves accuracy by 2-4\% (DMS Deep). The distribution of generated route lengths compared to the length of the experimental route is shown in Fig. S3-S4. Alternatively, providing the structure of the starting material (SM) allows to reduce the parameter count in half, and the resulting DMS Explorer (19M) model scores 4-7\% better on all Top-$K$ accuracies. Providing desired route length can further half parameter count (DMS Flash is only 10M) and add another 4-5\% in Top-1 and Top-10 accuracy. Notably, models that take extra information during training might still be used without that extra information, and would perform better than models of the same size trained without extra information. For example, a DMS Flash is trained with SM but might be evaluated without the structure of the SM, and the resulting performance (Tables~\ref{tab:n1_comparison}-\ref{tab:n5_comparison}) is 4-7\% better than that of equivalent 6-layer transformer trained without SM (entry 6x3-6x3 in Tables S4-S5). These results suggest a new training strategy if one is constrained by computational resources: extra input, derived directly from training data (DMS Flash is trained with the structure of the deepest node provided as SM) might serve as a learning aid, even if not used in route generation in production.

\subsubsection{Mixture of experts architecture reduces cost of generation}
Route generation time depends not only on the total number of parameters, but also on how those parameters are distributed. DMS Deep (38 M) utilizes a vanilla 36-layer decoder, meaning that every token prediction has to sequentially pass through 36 attention and multi-layer perceptron blocks. We find that redistributing the model parameters from the number of layers into the width of each layer by employing mixture-of-experts~\cite{shazeer2017outrageously,fedus2022switch,jiang2024mixtral}  based architecture (see Sec.~\ref{sec:model-cards} for details) allows a threefold reduction in the number of layers (and thus a threefold speedup): DMS Wide (38M) needs only 12-layer decoder and performs 1-2\% better than DMS Deep. Another important advantage of MoE-based models is that tokens passed through different active experts can be processed in parallel, which speeds up route generation even further.

Preliminary experiments showed that decoders with fewer than 6 layers perform significantly worse, making it impossible to train a size-comparable MoE alternative to DMS Flash. However, a 6-layer MoE trained with structures of SM (DMS Flex, 19M) similarly outperforms DMS Flash by 1-2\% at no increase in average runtime given that extra parameters are processed in parallel expert blocks. Just like DMS Flash, DMS Flex can be evaluated without SM and still have better Top-1 accuracy on both n$_1$/n$_5$ and better Top-10 on n$_5$ than existing methods. Interestingly, MoE-based models provide another dimension for flexibility: routes can be generated with a number of active experts different than that used during training. While DMS Flex (trained with 2 active experts) evaluated with 2 active experts (Duo mode) shows higher Top-K accuracy than if evaluated with 1 active expert (Mono mode), both result in comparable number of routes at different stages of post-processing (Tables S1-S2). As a result, one could consider predictions from Mono mode as a source of diversity.

Our experimentation with the increase in the number of experts, the number of encoder/decoder layers, or the feed-forward multiplier evaluated on small subsets of n$_1$ or n$_5$, show no significant improvement in performance (Tables S4-S8), which most likely means that the 20 M (with SM) or 40 M (without SM) are optimal given the current size of the training set.

\begin{figure}[htbp]
  \centering
  \includegraphics[width=\textwidth]{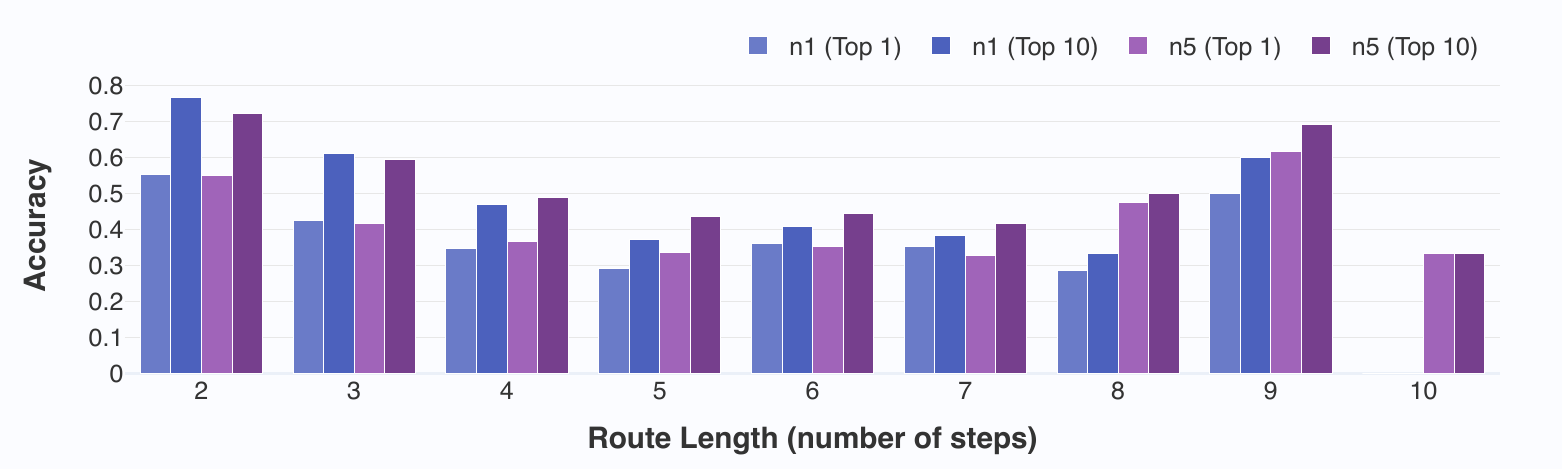}  
  \caption{Distribution of Top-1 and Top-10 accuracy of predictions with DMS-Flex (Duo) on test sets n$_1$ and n$_5$. There is only one route with length 10 in n$_1$, and DMS-Flex (Duo) does not predict it correctly. That route is reproduced by splitting it in half, as shown in Fig.~\ref{fig:n1step10}}.
  \label{fig:accuracy_dist}
\end{figure}

\subsubsection{Route quality as a function of route length}

Fig.~\ref{fig:accuracy_dist} shows the distribution of Top-1 and Top-10 accuracy over different route lengths on both n$_1$ and n$_5$. Given that 85\% of routes in the training partition have 4 or fewer steps Fig.~\ref{fig:length_distribution}), one would expect the accuracy to decrease dramatically with increasing route length. However, the performance on routes with 5-8 steps is comparable to that of shorter routes, and the performance on 9-step routes is even comparable to that of 2-step routes. Top-1 and Top-10 accuracies of other DMS models for different route lengths are given in Tables S9-S12.

\begin{figure}[htbp]
  \centering
  \includegraphics[width=\textwidth]{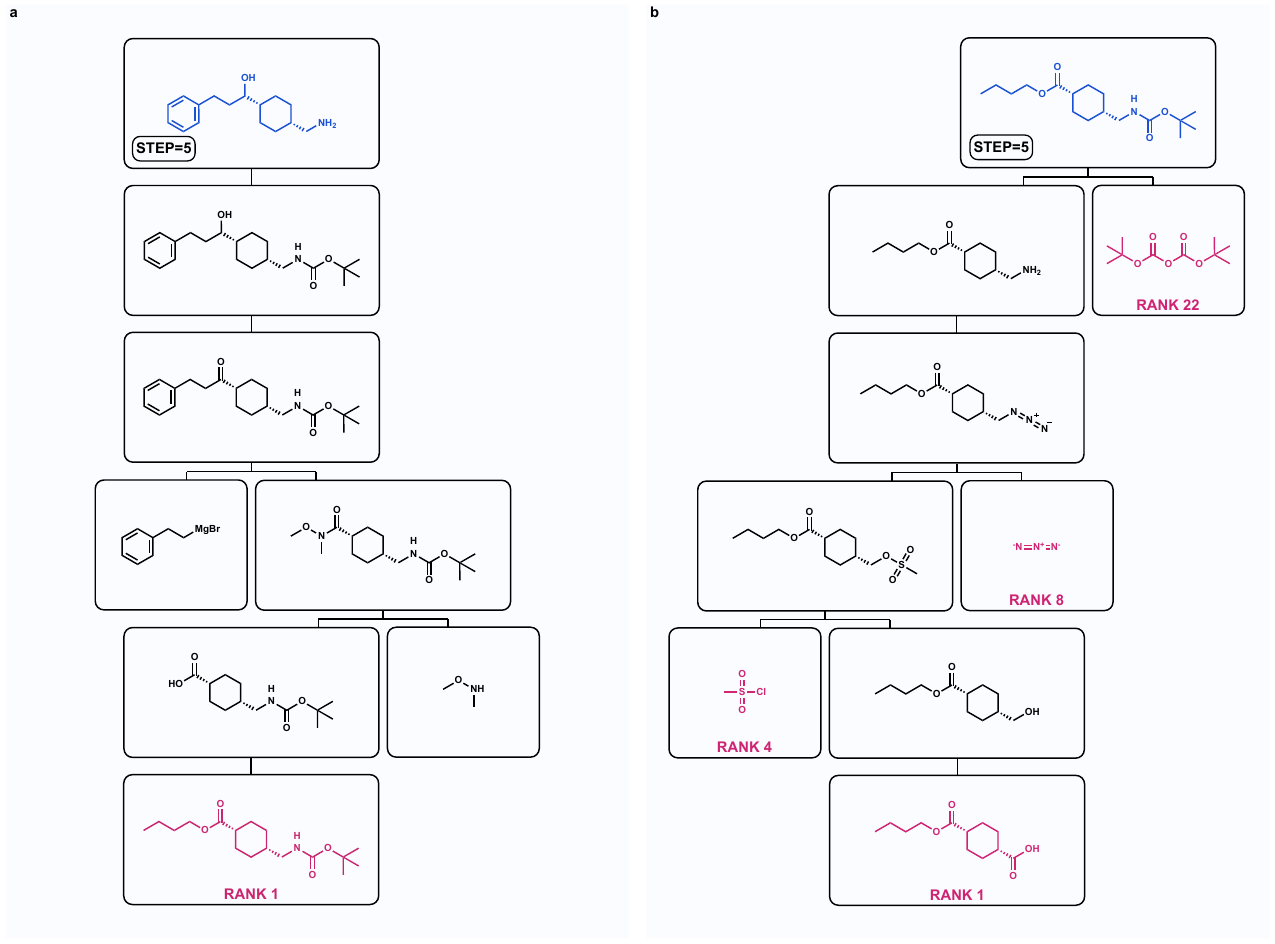}  
    \begin{subfigure}{0pt} % Empty subfigure for phantom label
        \phantomcaption
        \label{fig:n1step10_first}
    \end{subfigure}%
    
    \begin{subfigure}{0pt} % Empty subfigure for phantom label
        \phantomcaption
        \label{fig:n1step10_second}
    \end{subfigure}
    
  \caption{Separation of a 10-step route from set-n$_1$ into two 5-step routes. (a) Correct prediction from DMS-Flex (Duo) for the first half of the 10-step route with starting material information (red). (b) Correct prediction from DMS-Flex (Duo) for the second half of the 10-step route with starting material information.}
  \label{fig:n1step10}
\end{figure}

\subsubsection{Separation of Long Routes into Shorter Routes}

As seen in Fig.~\ref{fig:accuracy_dist}, DMS-Flex (Duo) does not find the single 10-step route in the n$_1$ set. However, this route can be correctly reproduced with rank 1 if split into two 5-step routes and the intermediate compound is provided as the starting material for the first 5 reactions (Fig.~\ref{fig:n1step10_first}) and a target compound for the last 5 reactions (Fig.~\ref{fig:n1step10_second}). While ideally the necessity to provide an intermediate compound should be avoided, this resembles the bidirectional search technique commonly employed in manual retrosynthetic analysis~\cite{corey_logic_1995}. In contrast, models trained without SM, such as DMS-Deep and DMS-Wide, can replicate the second 5-step sequence in Fig.~\ref{fig:n1step10_second} but not the first (Fig.~\ref{fig:n1step10_first}).

\subsubsection{Use Cases for Different DMS Model Variants} \label{use_cases}
The requirement to provide the starting material structure as an input to the model might, in theory, seem as a limitation, limiting the relevance of DMS Flash/Flex models. However, in practice, a chemist often can know the structure of a desired starting material because Corey's retrosynthetic framework allows finding at least one route to starting materials by performing one functional group transformation at a time. Such deterministically found routes will often be too long to be efficient in terms of yield (overall yield is the product of yields of individual steps). The art of organic synthesis lies in finding ways to perform several transformations simultaneously using a minimal number of protecting groups. As such, we believe our SM-based models provide value as an assistant to this task: finding better ways to get to the target compound from a starting material. The requirement to specify the number of steps as an input becomes a useful option in this context. Moreover, if the target compound is chiral (i.e., it has a mirror image which cannot be superimposed), even during manual retrosynthetic planning, chemists check if chiral centers can be obtained from chiral starting materials (such as naturally occurring amino acids) because developing reactions that create chiral centers in the desired conformation remains an active research area (and subject of 2001 and 2021 Nobel Prizes in Chemistry). An example of such a route is shown in Fig.~\ref{fig:daridorexant}. Alternatively, one can use more computationally expensive DMS Wide/Deep models to discover potential starting materials, and then use those structures as inputs to DMS Flash/Flex to generate suggestions for each starting material of interest. 

To lift the necessity to specify desired route length, a user also has a choice between running Flex/Wide models several times with different desired route lengths or simply running Explorer XL once. Of course, ideally one would prefer a single model that predicts the whole route with just from the structure of the target compound, without number of steps or starting material provided as an extra input. We believe better performance could be attainable in the future once bigger datasets become available. Finding ways to improve performance even with current dataset constitutes an important area of further research.

\begin{figure}[htbp]
  \centering
  \includegraphics[width=\textwidth]{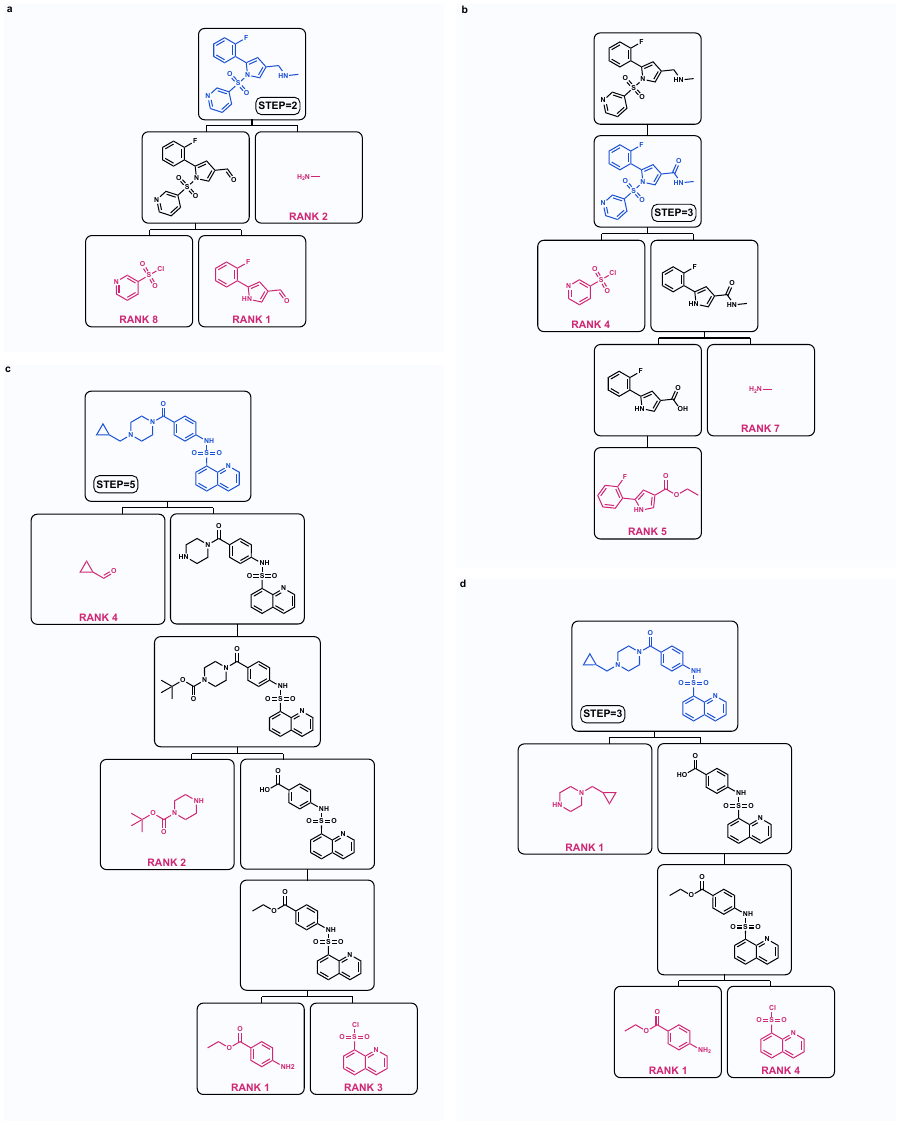}  
    \begin{subfigure}{0pt} % Empty subfigure for phantom label
        \phantomcaption
        \label{fig:vonoprazan_route1}
    \end{subfigure}%
    
    \begin{subfigure}{0pt} % Empty subfigure for phantom label
        \phantomcaption
        \label{fig:vonoprazan_route2}
    \end{subfigure}

    \begin{subfigure}{0pt} % Empty subfigure for phantom label
        \phantomcaption
        \label{fig:mitapivat_route1}
    \end{subfigure}

    \begin{subfigure}{0pt} % Empty subfigure for phantom label
        \phantomcaption
        \label{fig:mitapivat_route2}
    \end{subfigure}
    \vspace{-4em}
  \caption{Literature routes for Vonoprazan and Mitapivat correctly reproduced by DMS-Flex (Duo). Target compounds are in blue, starting materials that are given as inputs are colored in red. Ranks denote the rank of this route when the specified starting material is provided. (a) First literature route for Vonoprazan. The model predicts correctly no matter which starting material is given. (b) Second literature route for Vonoprazan. The model predicts the route correctly only when an immediate precursor to Vonoprazan is given as the target compound. (c) First literature route for Mitapivat. (d) Second literature route for Mitapivat.}
  \label{fig:vonomitaroutes}
\end{figure}

\subsection{Retrosynthetic planning of pharmaceutical compounds}

\begin{figure}[htbp]
  \centering
  \includegraphics[width=\textwidth]{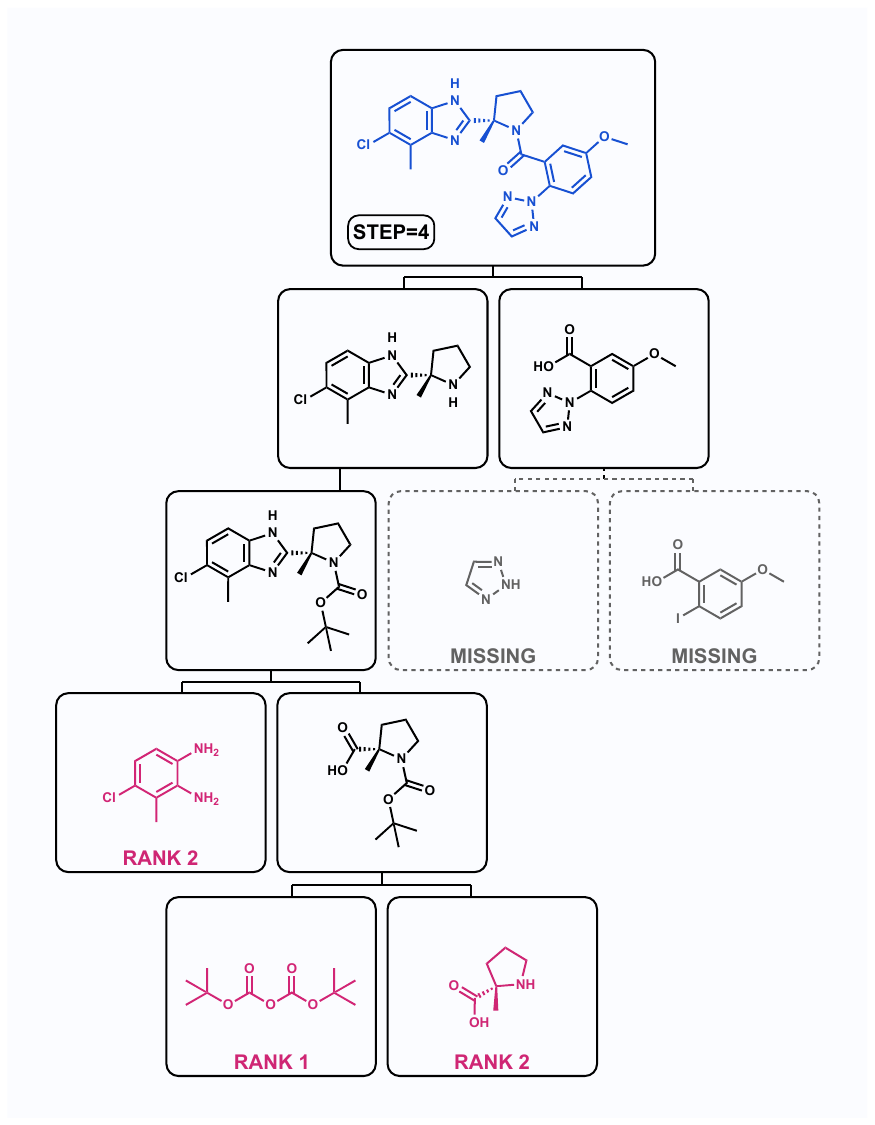}  
  \caption{Daridorexant literature multistep routes and predictions from DMS-Flex (Duo). The compounds that are missing from the prediction are in dotted boxes.}
  \label{fig:daridorexant}
\end{figure}

For another generalizability assessment of our model, we test it on three FDA-approved drugs: Vonoprazan, Mitapivat, and Daridorexant, which were used for evaluation by Xiong, et al.~\cite{xiong_improve_2023}. These drugs and their intermediates are absent from the training set. Vonoprazan, a potassium competitive acid blocker for Helicobacter pylori infections, was initially proposed with a 2-step route (Fig.~\ref{fig:vonoprazan_route1})~\cite{vonoprazan_patent_2007}. However, one of its starting materials is unstable, and the final reaction leads to significant byproducts. Subsequently, an improved 4-step route (Fig.~\ref{fig:vonoprazan_route2}) was introduced~\cite{yu_novel_2017}. Mitapivat, a pyruvate kinase activator for treating hemolytic anemia, has been associated with both a 5-step route (Fig.~\ref{fig:mitapivat_route1}) and a 3-step route (Fig.~\ref{fig:mitapivat_route2})~\cite{mitapivat_patent_2007, mitapivat_patent_2019}. Daridorexant, an orexin receptor antagonist for adult insomnia, is linked to a 4-step route (Fig.~\ref{fig:daridorexant})~\cite{daridorexant_patent_2023}.

Both vanilla transformers (DMS-Flash) and MoE models (DMS-Flex Duo) correctly reproduce Vonoprazan's first literature route (Fig.~\ref{fig:vonoprazan_route1}) regardless of the starting material provided, with high ranks. However, DMS-Flash struggles to find the first transformation (counting from the target compound) in the second route. This inability could be attributed to the fact that the reduction of amide carbonyl (the step leading to Vonoprazan in second route) is less frequently represented in the PaRoutes dataset than the much more common reductive amination (the step leading to Vonoprazan in first route). As a result, most routes predicted by DMS-Flash start with a reductive amination as in Fig.~\ref{fig:vonoprazan_route1}. However, if we provide the precursor to Vonoprazan in the second route as the target compound, the model correctly reproduces the remaining route (Fig.~\ref{fig:vonoprazan_route2}). This underscores that the performance of our models relies upon sufficient representation of transformations in the training set. This limitation, however, should naturally disappear once larger multistep datasets become available.

\begin{figure}[htbp]
  \centering
  \includegraphics[width=\textwidth]{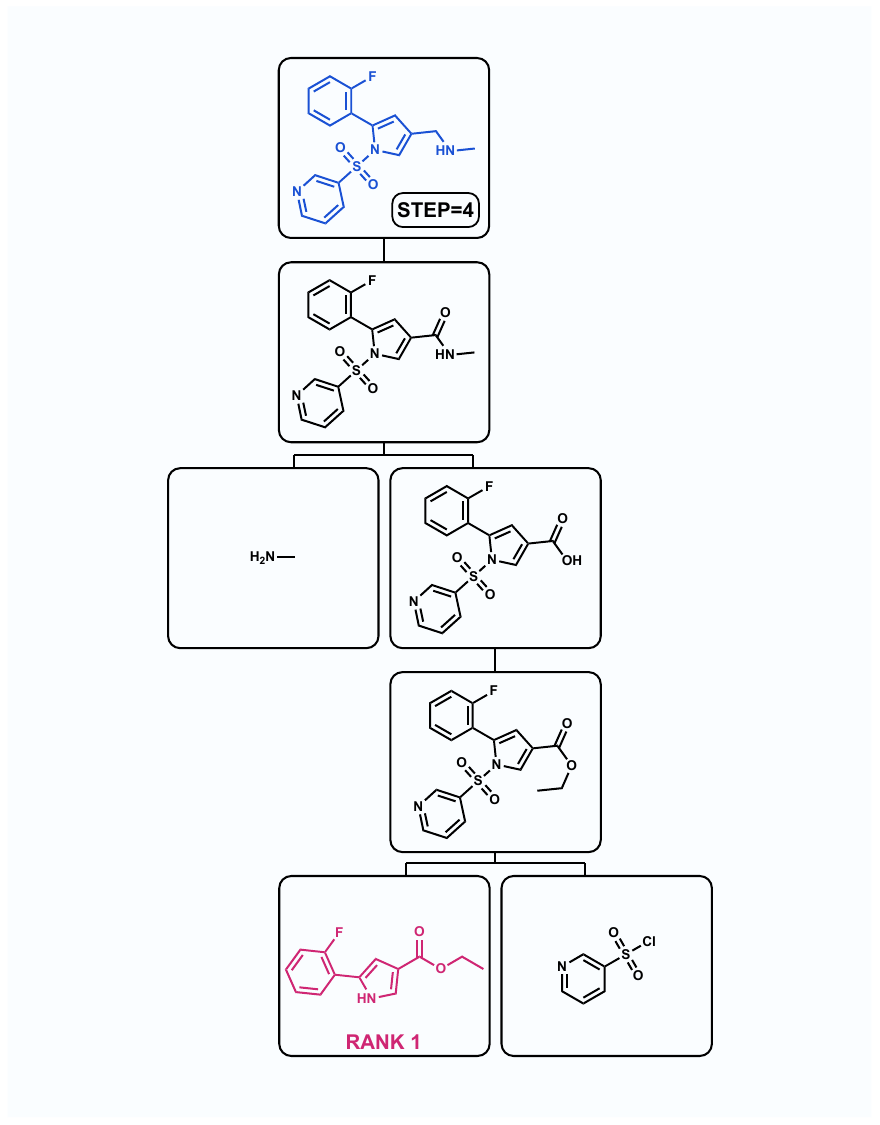}  
  \caption{A variation of second Vonoprazan route predicted by DMS-Flex (Duo). The correct route is shown in Fig.~\ref{fig:vonoprazan_route2} and differs by the order of the last steps: tosyl group is introduced after amidation, not before.}
  \label{fig:vonoprazan_wrong_order}
\end{figure}

Inability of DMS-Flash to suggest reduction of amide carbonyl in Fig.~\ref{fig:vonoprazan_route2} was the motivation for the creation of mixture-of-experts (MoE) models. We conjectured that different experts might specialize in different classes of transformations, and even if one expert dominates the generations, one could still manually force the output from a different expert. Fortunately, no such manual interventions were needed as DMS-Flex (Duo) predicted the route almost identical to the second transformation as the Top-1 candidate (Fig. \ref{fig:vonoprazan_wrong_order}). The only difference from the correct route is the order of transformations: in the correct route the tosyl group is introduced after amidation with methylamine, not before.

DMS-Flex (Duo) correctly reproduces both routes for Mitapivat (Fig.~\ref{fig:mitapivat_route1} and Fig.~\ref{fig:mitapivat_route2}). In contrast, the route for Daridorexant (Fig.~\ref{fig:daridorexant}) is only partially predicted: our model does not split the right child of the root node into the two compounds in the dotted boxes. All other intermediates and starting materials are predicted correctly. The reason for this behavior is not well understood, but it can be rationalized by looking at the distribution of the number of leaves at root nodes: 73\% of root nodes (target compounds) have at least 1 leaf (see Fig. S5). This dependence on the distributions of the training set comprises an important limitation of our models; however, it should also be resolved once bigger datasets become available. Admittedly, this issue would not occur with current methods based on single-step prediction because the tree search framework would not halt at compounds outside the stock compound set before reaching the maximum iterations. Ensuring that multistep-first models reproduce that aspect of SSR methods constitutes a direction for future research.

This example, however, does not imply that the model struggles to predict routes with convergent branches. If one defines a convergent route as the one that has at least one node (compound) with at least two children that are not leaves, and repeats the analysis, no significant differences in the number of routes found (Fig. S1-S2) or Top-K accuracies (Fig. S7-S8) is observed despite significant variation in the number of routes that are convergent among training routes, n$_1$, and n$_5$ (Fig. S6).

\section{Conclusion}
In this work, we introduce the first-of-its-kind inherently multistep approach for multistep retrosynthesis prediction. By predicting the entire route as a single string, our approach bypasses the need for iterative single-step predictions and exponential search space traversal. We present a family of transformer-based DirectMultiStep (DMS) models which can predict correct routes just from the structure of the target compound, but can also accommodate specific conditions such as the desired number of steps and starting materials, enabling efficient route planning.

Our flagman DMS-Flex (Duo) model, incorporating mixture-of-experts approach, outperforms state-of-the-art methods on the PaRoutes dataset, achieving a 2.5x improvement in Top-1 accuracy on an n$_1$ test set and a 3.9x improvement on the n$_5$ test set. We demonstrate the quality of predicted routes by testing our models on several FDA-approved drugs. Those predictions also highlight limitations caused by the suboptimal diversity of the current training set, particularly for less common reaction types. Those limitations are partially mitigated by employing mixture-of-experts techniques, making them a promising candidate for further research in retrosynthesis prediction and applications of generative models to chemistry in general.

We believe our work demonstrates the superiority of multistep-first approach to full route prediction and inspires further research into methods that reduce the exponential complexity of iterative application of single-step based methods. We are grateful to the authors of PaRoutes for the development and release of the first multistep prediction benchmark, and we hope more benchmarks based on experimentally verified routes become available. Future research may focus on improvement of performance on longer routes (e.g. eliminating the necessity to split longer routes) and (or) incorporation of additional constraints (such as cost) on the route prediction. Overall, this work demonstrates the potential of transformer-based models for efficient and scalable computer-aided synthesis planning. We release the code used for training and route prediction as a public \href{https://github.com/batistagroup/DirectMultiStep}{GitHub Repo} under MIT License. Our models are also available with a Web GUI at \href{https://models.batistalab.com}{models.batistalab.com}.

\section{More Computational Details}
\subsection{Data Curation and Preparation}

A second version of the PaRoutes~\cite{genheden_paroutes_2022} dataset containing 450k routes is stripped of all metadata and stored as recursive dictionaries representing multistep routes. The two evaluation sets, n$_1$ and n$_5$, are processed similarly. A training partition is created by removing all repetitions (some routes were represented multiple times with different metadata) and permutations (swapping left and right subtrees) of routes in n$_1$ and n$_5$ from the full dataset, resulting in 163k routes. The training dataset is augmented by adding 2 permutations for each route. To train the DMS model which takes starting material as an input, we find all starting materials (leaves) for each tree and store a combination of the target compound with each starting material as a separate entry. As a result, DMS with SM is trained on 1 340 243 inputs, and the DMS without SM on 432 684 inputs. All SMILES strings are tokenized by treating each character as a token, and the string representation of the multistep route is tokenized similarly, treating delimiters of the tree (\verb+`smiles'+, \verb+`children'+, \verb+,+, \verb+[+, \verb+]+, \verb+{+, and \verb+}+) as separate tokens. The final vocabulary size is 52 (including start, end, and padding tokens), the largest multistep route has 1074 tokens, while the largest target compound and starting material have 145 and 135 tokens, respectively.

\subsection{Model Architecture} \label{sec:model-cards}
We present a family of DirectMultiStep (DMS) models with varying parameter counts and architectural details. DMS-Flash and DMS-Deep employ the classical transformer encoder-decoder architecture~\cite{vaswani_attention_2017}, whereas DMS-Flex and DMS-Wide incorporate sparsely-gated mixture-of-experts layers~\cite{shazeer2017outrageously,fedus2022switch,jiang2024mixtral}. To create MoE model, one replaces the 2-layer MLP that follows every attention block with an ensemble of 2-layer MLPs (called experts). A preceding router determines (for each input) the $n$-most relevant (active) experts, and passes the input through each of them. Output of each active expert is added with weights determined by the router. Notably, the number of experts active during inference does not have to be equal to the number of experts active during training, although the latter tends to be optimal. Major parameters choices are given in model card (Table ~\ref{tab:model_specs}). All models use GeLU activations~\cite{hendrycks2023gaussian}.

\begin{table}[htbp]
	\centering
	\caption{Specifications of architecture for different model variants.}
	\label{tab:model_specs}
	\begin{tabular}{lccccccc}
		\toprule
		DMS Model    & Encoder& Decoder& Hidden    & FF         & Total   & \multicolumn{2}{c}{Active Experts} \\
		Variant      & Layers & Layers & Dimension & Multiplier$^a$ & Experts & Training & Generation \\
		\midrule
		Flash        & 6      & 6      & 256       & 3          & 1       & 1        & 1          \\
            Explorer     & 6      & 6      & 256       & 3          & 3       & 2        & 2          \\
            Explorer XL  & 8      & 24     & 256       & 3          & 3       & 2        & 2          \\
		Flex (Mono)  & 6      & 6      & 256       & 3          & 3       & 2        & 1          \\
		Flex (Duo)   & 6      & 6      & 256       & 3          & 3       & 2        & 2          \\
		Deep         & 12     & 36     & 256       & 3          & 1       & 1        & 1          \\
		Wide         & 12     & 12     & 256       & 3          & 3       & 2        & 2          \\
		\bottomrule
	\end{tabular}\\
 \footnotesize{$^a$ an integer by which the hidden dimension is multiplied in feed-forward layers after attention.}
\end{table}

Intermediate experiments with evaluations on subsets of n$_1$ and n$_5$ with 50 and 500 routes reveal no benefit in increasing the number of hidden dimensions to 512 (Tables S4-S6), which could be rationalized by noticing that 256 is already almost 5x the vocabulary size. Although a feed-forward multiplier of four is more common in NLP models, we find no significant improvement in performance as compared to the feed-forward multiplier of three (Tables S7-S8). In addition, the number of encoder layers does not have to be the same as the number of decoder layers, and models with 6 encoder and 12 decoder layers (trained with SM) performed almost as good as the models with 12 encoder and decoder layers. For MoE models, we find no significant improvement upon increase of number of experts to four, and models trained with two active experts outperform those trained with one.

\subsection{Training}
During training, a 10\% dropout rate is used in embedding, attention, and feed-forward layers. The AdamW optimizer~\cite{kingma2014adam,2019adamw} is used with a learning rate scheduler that warms up from 0 to $3.5\times 10^{-4}$ for DMS-Flex and DMS-Wide, $3\times 10^{-4}$ for DMS-Flash, and $1\times 10^{-4}$ for DMS-Deep during the first 10\% of training steps and undergoes cosine decay to $10\%$ of the maximal learning rate over 20 epochs. The optimal initial learning rate for each model was chosen by starting from 5e-4 and monitoring for gradient explosion (in which case the learning rate was decreased by 0.5e-4). Gradients are clipped to 1.0. In addition, each token in the input to the encoders has a 5\% chance of being masked. During hyperparameter optimization, the model was validated on a 5\% split from the training routes (which exclude n$_1$ and n$_5$ routes).

\section{Data and Software Availability}
The reaction route dataset is from the 2.0 version of PaRoutes in their GitHub repository~\cite{genheden_paroutes_2022}, available under the Apache License 2.0. Code to process dataset, implementation of model architecture, code for training, generation, and evaluation are available under MIT License at \url{https://github.com/batistagroup/DirectMultiStep}. A web interface is also available at \href{https://models.batistalab.com}{models.batistalab.com}.

\section*{Author Contributions}
YS and AM contributed equally to this work. YS conceived the idea. YS and HL conducted preliminary model training and testing. AM introduced the mixture of experts (MoE) approach. AM and YS carried out additional model training and testing and wrote the manuscript. VSB provided guidance for the research. All authors have given approval to the final version of the paper.

\section*{Conflict of Interest}
The authors declare no conflict of interest.

\section{Acknowledgement}

The authors acknowledge a generous allocation of high-performance computing time from NERSC. The development of the methodology was supported by the NSF CCI grant (VSB, Award Number 2124511).

% \clearpage
\medskip
{
\small
\bibliography{reference}
}

%%%%%%%%%%%%%%%%%%%%%%%%%%%%%%%%%%%%%%%%%%%%%%%%%%%%%%%%%%%%
\newpage
\renewcommand{\thetable}{S\arabic{table}} % Prefix table numbers with 'S'
\renewcommand{\thefigure}{S\arabic{figure}} % Prefix figure numbers with 'S'
\setcounter{table}{0} % Reset table counter
\setcounter{figure}{0} % Reset figure counter

\section*{Supporting Information}

\begin{table}[htbp]
	\caption{Number of predicted routes in n$_1$ that pass post-processing filters.}
	\centering
	\begin{tabular}{lcccc}
		\toprule
		Model           & Valid$^a$ & Correct SM$^{b, f}$ & SMs In Stock$^c$  & Correct and In Stock$^{d, f}$\\
		\midrule
            Explorer-XL$^e$ & 9998  &      & 8008 &            \\
		Deep$^e$        & 9998  &      & 7957 &            \\
            Explorer        & 9997  & 9923 & 8681 & 8555       \\
            Explorer$^e$    & 9995  &      & 7418 &            \\
		Flash           & 9998  & 9942 & 8814 & 8725       \\
            Flash$^e$       & 9994  &      & 7660 &            \\
		Wide$^e$        & 9996  &      & 8089 &            \\
		Flex (Mono)     & 9941  & 9545 & 8338 & 7948       \\
		Flex (Duo)      & 9995  & 9866 & 8828 & 8649       \\
		Flex (Duo)$^e$  & 9994  &      & 7692 &            \\
		\bottomrule
	\end{tabular}
\label{table:n1_search_performance}
\begin{minipage}{\textwidth}
\raggedright
\footnotesize{$^a$At least one route with all valid SMILES (including target, intermediates, and SMs).}
\\\footnotesize{$^b$At least one route with the correct conditioned SMs (and with all valid SMILES).}
\\\footnotesize{$^c$At least one route with all SMs in the n$_1$ stock set (and with all valid SMILES).}
\\\footnotesize{$^d$At least one route with the correct conditioned SMs and with all SMs in the n$_1$ stock set (and with all valid SMILES).}
\\\footnotesize{$^e$SMs not provided during inference.}
\\\footnotesize{$^f$Blank indicates not relevant (models without SM conditions).}
\end{minipage}
\end{table}

\begin{table}[htbp]
	\caption{Number of predicted routes in n$_5$ that pass post-processing filters.}
	\centering
	\begin{tabular}{lcccc}
		\toprule
		Model           & Valid$^a$ & Correct SM$^{b, f}$ & SMs In Stock$^c$  & Correct and In Stock$^{d, f}$\\
		\midrule
            Explorer-XL$^e$ &10000  &      & 7904 &            \\
		Deep$^e$        & 9997  &      & 7830 &            \\
            Explorer        & 9997  & 9903 & 8439 & 8281       \\
            Explorer$^e$    & 9998  &      & 7227 &            \\
		Flash           & 9991  & 9888 & 8646 & 8487       \\
            Flash$^e$       & 9991  &      & 7576 &            \\
	    Wide$^e$        & 9997  &      & 7950 &            \\
		Flex (Mono)     & 9932  & 9374 & 8037 & 7528       \\
		Flex (Duo)      & 9993  & 9808 & 8627 & 8382       \\
		Flex (Duo)$^e$  & 9990  &      & 7605 &            \\
		\bottomrule
	\end{tabular}
\label{table:n5_search_performance}
\begin{minipage}{\textwidth}
\raggedright
\footnotesize{$^a$At least one route with all valid SMILES (including target, intermediates, and SMs).}
\\\footnotesize{$^b$At least one route with the correct conditioned SMs (and with all valid SMILES).}
\\\footnotesize{$^c$At least one route with all SMs in the n$_5$ stock set (and with all valid SMILES).}
\\\footnotesize{$^d$At least one route with the correct conditioned SMs and with all SMs in the n$_5$ stock set (and with all valid SMILES).}
\\\footnotesize{$^e$SMs not provided during inference.}
\\\footnotesize{$^f$Blank indicates not relevant (models without SM conditions).}
\end{minipage}
\end{table}

\begin{figure}[htbp]
  \centering
  \includegraphics[width=\textwidth]{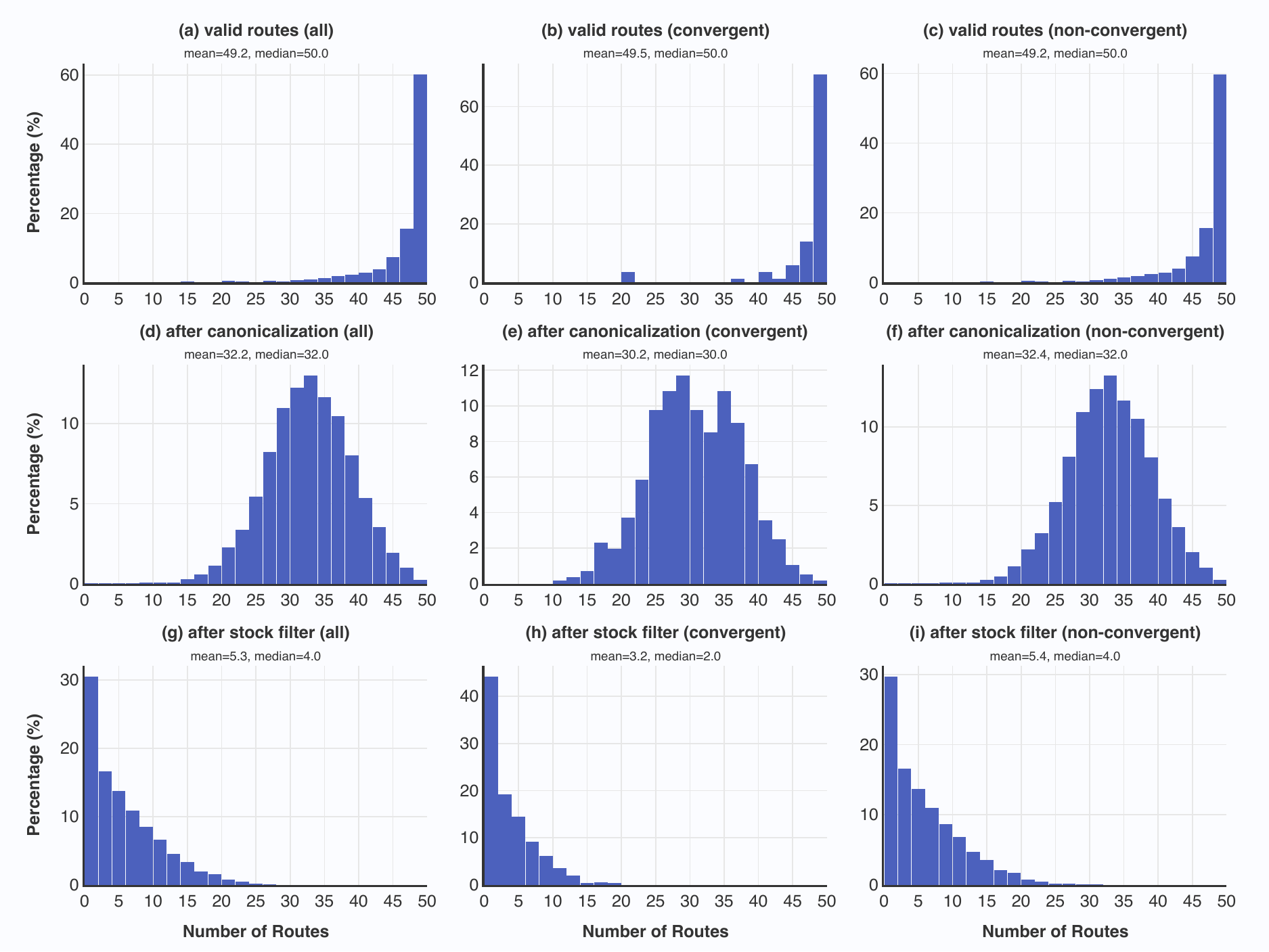}  
  \caption{Distribution of the average number of routes predicted by DMS Explorer XL for targets in n$_1$ test set (a-c) after checking for the validity of smiles, (d-f) after canonicalization of SMILES and removing repetitions (considering permutations of children), and (g-i) after filtering for routes that have all reactants in stock. Distribution is shown separately for all targets (a, d, g), for targets for which an experimental route is convergent (b, e, h) and for targets for which an experimenal route is non-convergent (c, f, i). A route is considered convergent if any node has at least two children that are not leaves.}
  \label{fig:n1_processing_stages}
\end{figure}

\begin{figure}[htbp]
  \centering
  \includegraphics[width=\textwidth]{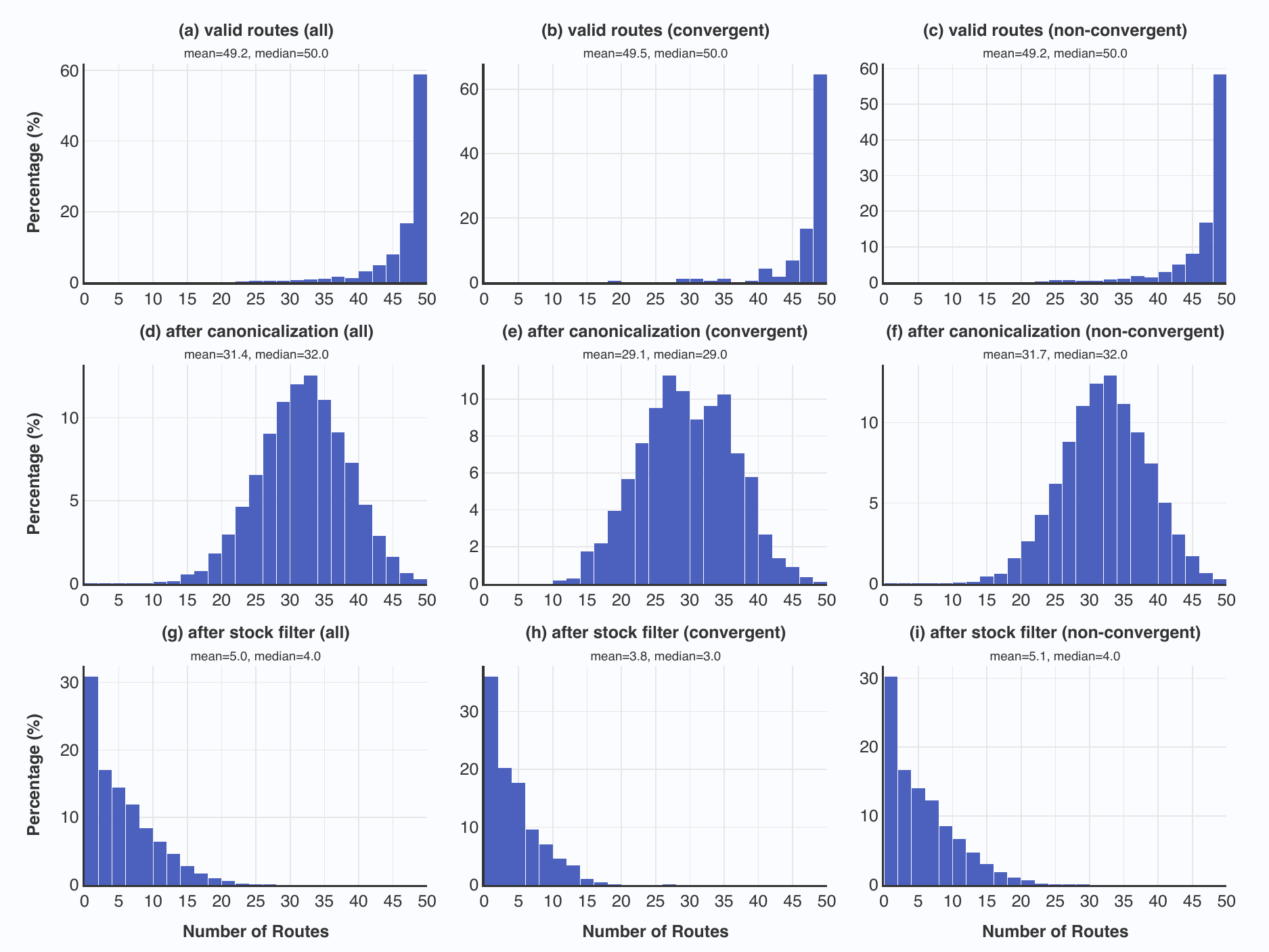}  
  \caption{Distribution of the average number of routes predicted by DMS Explorer XL for targets in n$_5$ test set (a-c) after checking for the validity of smiles, (d-f) after canonicalization of SMILES and removing repetitions (considering permutations of children), and (g-i) after filtering for routes that have all reactants in stock. Distribution is shown separately for all targets (a, d, g), for targets for which an experimental route is convergent (b, e, h) and for targets for which an experimenal route is non-convergent (c, f, i). A route is considered convergent if any node has at least two children that are not leaves.}
  \label{fig:n5_processing_stages}
\end{figure}

\begin{table}[htbp]
	\caption{Search performance for DirectMultiStep models on USPTO-190~\cite{chen_retro_2020}}
	\centering
	\begin{tabular}{lccc}
		\toprule
		Method$^{a}$                          & Stock Set$^{b}$      & Solved Rate$^{c}$ & Run Time (s)$^d$ \\
		\midrule
		DMS-Explorer-XL                       & eMolecules Screening & 33.2\% (31.5\%)      & 21.9             \\
		MCTS~\cite{chen_retro_2020}           & eMolecules Screening & 33.7\%      & 370.5            \\
		DMS-Flash$^{e,f}$              & eMolecules Screening & 60.5\% (55.2\%)      & 31.3             \\
		DMS-Wide$^e$                   & eMolecules Screening & 63.7\% (56.6\%)      & 105.8            \\
		Retro*~\cite{chen_retro_2020}         & eMolecules Screening & 86.8\%      & 156.6            \\
		RetroGraph~\cite{xie2022retrograph}   & eMolecules Screening & 99.5\%      & NA               \\
		\midrule
		DMS-Explorer-XL                       & Buyables             & 27.9\% (28.0\%)      & 21.9             \\
		Original + MCTS~\cite{roh2025higher}  & Buyables             & 46.3\%      & NA               \\
		DMS-Flash$^{e,f}$              & Buyables             & 55.3\% (51.0\%)      & 31.3             \\
		DMS-Wide$^e$                   & Buyables             & 56.8\% (51.7\%)      & 105.8            \\
		Higherlev + MCTS~\cite{roh2025higher} & Buyables             & 73.7\%      & NA               \\
		\bottomrule
	\end{tabular}
	\label{table:pto190_search_performance}
	\begin{minipage}{\textwidth}
		\raggedright
		\footnotesize{$^a$ All DMS models are run with a beam size of 50 on a single NVIDIA A100 GPU with half-precision floating point inference (fp16).}  \\
		\footnotesize{$^b$ eMolecules Screening is from Chen et al.~\cite{chen_retro_2020} and contains 23.1M screening compounds from eMolecules as of 2019 (2019-11-01). Buyables is from Roh et al.~\cite{roh2025higher} and includes 0.329M buyable building blocks from eMolecules, Sigma-Aldrich, Mcule, ChemBridge Hit2Lead, and WuXi LabNetwork.} \\
		\footnotesize{$^c$ 47 of the targets from USPTO-190 are present as non-leaf nodes in PaRoutes training set. The solved rate on the remaining 143 targets is reported in parentheses.} \\
		\footnotesize{$^d$ Averages over all targets. NA indicates that the run time is not available in the original publication.}  \\
		\footnotesize{$^e$ Uses step counts from 2 to 8 (total of 7 DMS model runs).}\\
		\footnotesize{$^f$ SMs not provided for DMS-Flash.}\\
		
	\end{minipage}
\end{table}

\begin{figure}[htbp]
  \centering
  \includegraphics[width=\textwidth]{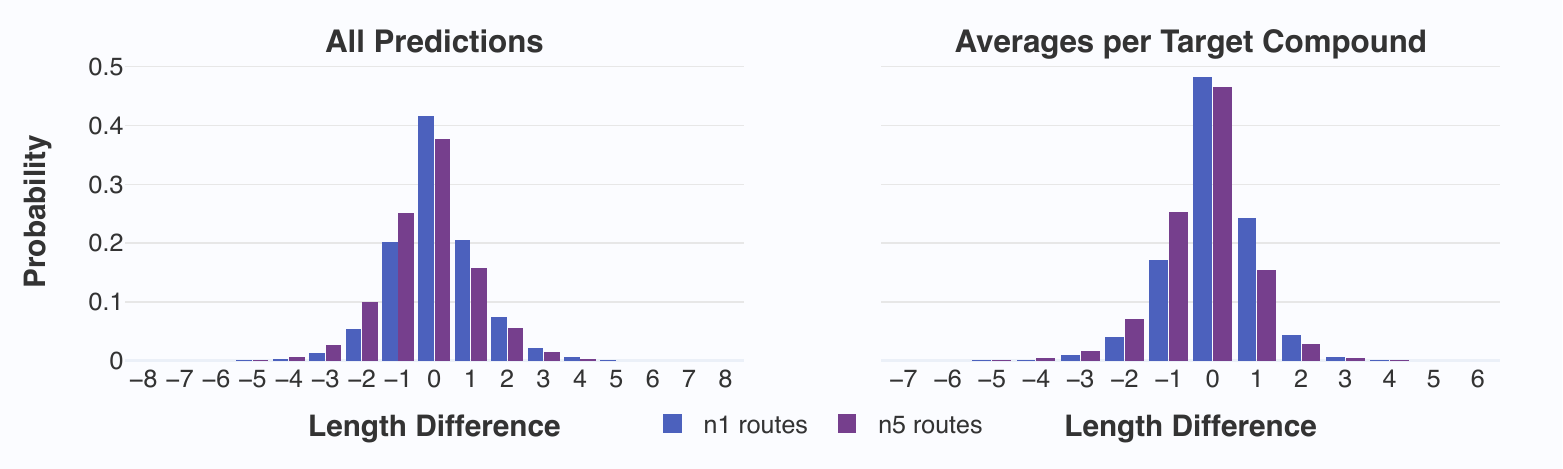}  
  \caption{Lengths of valid routes predicted by DMS-Explorer for targets in n$_1$ and n$_5$ compared to the length of the correct (experimental) route. Left: distribution of differences in lengths for all predictions for all targets. Right: distribution of average (among all predictions for a single target) differences.}
  \label{fig:explorer_dev_hist}
\end{figure}

\begin{figure}[htbp]
  \centering
  \includegraphics[width=\textwidth]{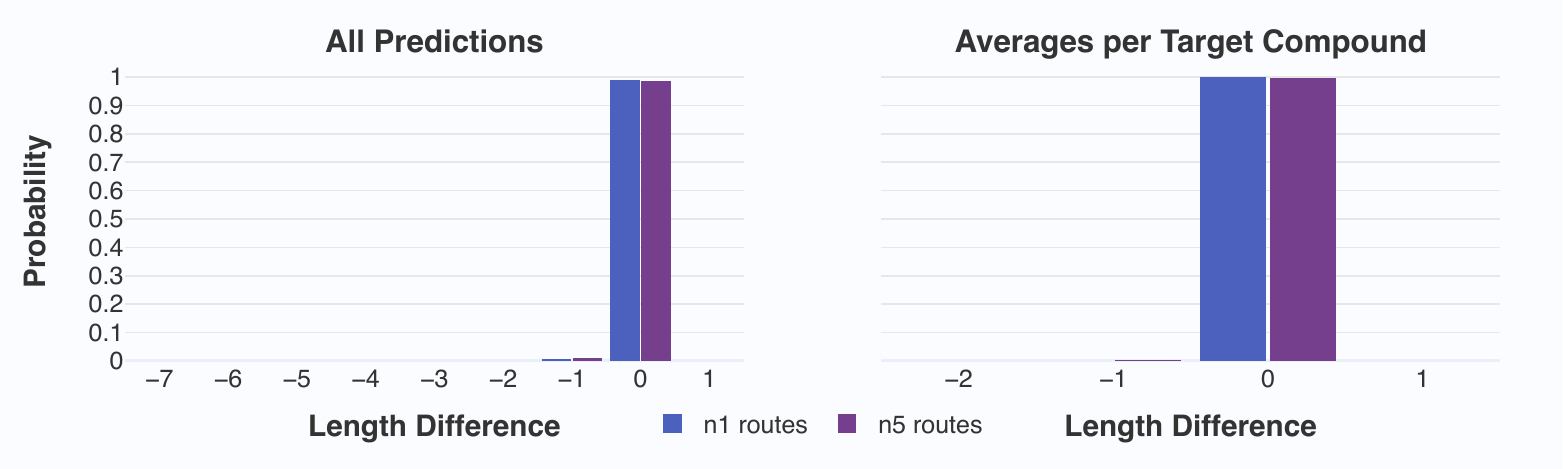}  
  \caption{Lengths of valid routes predicted by DMS-Flex (Duo) for targets in n$_1$ and n$_5$ compared to the length of the correct (experimental) route. Left: distribution of differences in lengths for all predictions for all targets. Right: distribution of average (among all predictions for a single target) differences.}
  \label{fig:flex_duo_dev_hist}
\end{figure}

\begin{table}[htbp]
	\caption{Top-$K$ accuracy on subset$^a$ of 500 routes from test set-n$_1$.}
	\centering
	\begin{tabular}{llcccllllll}
		\toprule
		       &      &        & SM      & SM     &       &       &       &       &       &        \\
		Models    & Size & Steps  & (Train) & (Gen.) & Top-1 & Top-2 & Top-3 & Top-4 & Top-5 & Top-10 \\
		\midrule 
		Flash   & 10M  & \cmark & \cmark  & \cmark & 0.402 & 0.476 & 0.522 & 0.550 & 0.564 & 0.582  \\
		6x4-12x4$^b$  & 18M  & \cmark & \cmark  & \cmark & 0.446 & 0.552 & 0.598 & 0.616 & 0.624 & 0.640  \\
		12x3-12x3 & 19M  & \cmark & \cmark  & \cmark & 0.452 & 0.570 & 0.596 & 0.618 & 0.626 & 0.636  \\
		12x4-12x4 & 22M  & \cmark & \cmark  & \cmark & 0.452 & 0.552 & 0.588 & 0.610 & 0.620 & 0.642  \\
		6x3-6x3   & 10M  & \cmark & \xmark  & \xmark & 0.292 & 0.362 & 0.384 & 0.388 & 0.396 & 0.402  \\
		\bottomrule
	\end{tabular}
\label{tab:si_n1_topk_500}
 \raggedright
 \\\footnotesize{$^a$The subset was chosen randomly with a seed of 42. Flash results are provided to establish how the performance on these 500 routes relates to the performance on full test set.}
 \\\footnotesize{$^b$The model names follow NxF-KxF format, where N is the number of encoder layers, K is the number of decoder layers, and F is the feed-forward multiplier}
\end{table}

\begin{table}[htbp]
	\caption{Top-$K$ accuracy on subset$^a$ of 500 routes from test set-n$_5$.}
	\centering
	\begin{tabular}{llcccllllll}
		\toprule
		       &      &        & SM      & SM     &       &       &       &       &       &        \\
		Models    & Size & Steps  & (Train) & (Gen.) & Top-1 & Top-2 & Top-3 & Top-4 & Top-5 & Top-10 \\
		\midrule 
		Flash     & 10M  & \cmark & \cmark  & \cmark & 0.374 & 0.442 & 0.474 & 0.484 & 0.498 & 0.516  \\
		6x4-12x4$^b$  & 18M  & \cmark & \cmark  & \cmark & 0.400 & 0.464 & 0.498 & 0.520 & 0.530 & 0.548  \\
		12x3-12x3 & 19M  & \cmark & \cmark  & \cmark & 0.406 & 0.472 & 0.496 & 0.518 & 0.524 & 0.538  \\
		12x4-12x4 & 22M  & \cmark & \cmark  & \cmark & 0.370 & 0.448 & 0.472 & 0.498 & 0.508 & 0.526  \\
		6x3-6x3   & 10M  & \cmark & \xmark  & \xmark & 0.222 & 0.264 & 0.282 & 0.288 & 0.290 & 0.298  \\
		\bottomrule
	\end{tabular}
\label{tab:si_n5_topk_500}
\raggedright
 \\\footnotesize{$^a$The subset was chosen randomly with a seed of 42. Flash results are provided to establish how the performance on these 500 routes relates to the performance on full test set.}
 \\\footnotesize{$^b$The model names follow NxF-KxF format, where N is the number of encoder layers, K is the number of decoder layers, and F is the feed-forward multiplier}
\end{table}

\begin{table}[htbp]
	\centering
	\caption{Specifications of architecture for different model variants.}
	\begin{tabular}{lccccccc}
		\toprule
		DMS Model    & Encoder& Decoder& Hidden    & FF         & Total   & \multicolumn{2}{c}{Active Experts} \\
		Variant & Layers & Layers & Dimension & Multiplier$^a$ & Experts & Training & Generation \\
		\midrule
		Van-19  & 6      & 6      & 256       & 3              & 1       & 1        & 1          \\
		Van-60  & 8      & 8      & 512       & 4              & 1       & 1        & 1          \\
		MoE-1/3 & 6      & 6      & 256       & 3              & 3       & 1        & 1          \\
		\bottomrule
	\end{tabular}\\
\label{tab:si_architecture}
\raggedright
\footnotesize{$^a$ an integer by which the hidden dimension is multiplied in feed-forward layers after attention.}
\end{table}

\begin{table}[htbp]
	\caption{Top-$K$ accuracy on route test set-n$_1$ (10 000 routes).}
	\centering
	\begin{tabular}{llcccllllll}
		\toprule
		DMS     &      &        & SM      & SM     &       &       &       &       &       &        \\
		Models  & Size & Steps  & (Train) & (Gen.) & Top-1 & Top-2 & Top-3 & Top-4 & Top-5 & Top-10 \\
		\midrule 
		Van-19  & 19M  & \cmark & \cmark  & \cmark & 0.44  & 0.53  & 0.57  & 0.59  & 0.60  & 0.63   \\
		MoE-1/3 & 38M  & \cmark & \xmark  & \xmark & 0.40  & 0.47  & 0.50  & 0.51  & 0.52  & 0.53   \\
		Van-60  & 60M  & \cmark & \cmark  & \cmark & 0.38  & 0.46  & 0.48  & 0.50  & 0.51  & 0.52   \\
		Van-60  & 60M  & \cmark & \xmark  & \xmark & 0.36  & 0.42  & 0.44  & 0.45  & 0.45  & 0.46   \\
		\bottomrule
	\end{tabular}
\label{tab:si_n1_topk}
\end{table}

\begin{table}[htbp]
	\caption{Top-$K$ accuracy on route test set-n$_5$ (10 000 routes).}
	\centering
	\begin{tabular}{llcccllllll}
		\toprule
		DMS         &      &        & SM      & SM     &            &            &            &            &            &            \\
		Models      & Size & Steps  & (Train) & (Gen.) & Top-1      & Top-2      & Top-3      & Top-4      & Top-5      & Top-10     \\
		\midrule 
		Van-19          & 19M  &\cmark & \cmark  & \cmark & 0.39  & 0.47  & 0.50  & 0.52  & 0.53  & 0.56      \\
		MoE-1/3 & 38M  & \cmark & \xmark  & \xmark & 0.36  & 0.42  & 0.45  & 0.46  & 0.46  & 0.48   \\
		Van-60 & 60M & \cmark & \cmark & \cmark     & 0.38  & 0.46  & 0.48 & 0.50 & 0.51   & 0.52   \\
		Van-60 & 60M & \cmark &\xmark &\xmark    & 0.36           & 0.42  & 0.44 & 0.45 & 0.45   & 0.46   \\
		\bottomrule
	\end{tabular}
\label{tab:si_n5_topk}
\end{table}

\begin{table}[htbp]
\caption{Top-1 Accuracy on n$_1$ routes split by route length}
	\centering
	\begin{tabular}{lccccccccc}
		\toprule
		Model            & 2     & 3     & 4     & 5     & 6     & 7     & 8     & 9     & 10    \\
		\midrule
            Explorer-XL$^{a, b}$  & 0.402 & 0.304 & 0.271 & 0.259 & 0.296 & 0.277 & 0.286 & 0.600 & 0.000 \\
		Deep$^b$         & 0.497 & 0.395 & 0.338 & 0.268 & 0.272 & 0.262 & 0.286 & 0.600 & 0.000 \\
		Explorer$^a$     & 0.480 & 0.346 & 0.282 & 0.266 & 0.305 & 0.323 & 0.286 & 0.500 & 0.000 \\
		Flash            & 0.538 & 0.410 & 0.323 & 0.273 & 0.329 & 0.308 & 0.286 & 0.400 & 0.000 \\
		Wide$^b$         & 0.511 & 0.416 & 0.344 & 0.308 & 0.329 & 0.292 & 0.286 & 0.600 & 0.000 \\
		Flex (Mono)      & 0.456 & 0.318 & 0.226 & 0.193 & 0.183 & 0.154 & 0.143 & 0.300 & 0.000 \\
		Flex (Duo)       & 0.554 & 0.426 & 0.348 & 0.292 & 0.362 & 0.354 & 0.286 & 0.500 & 0.000 \\
		Flex (Duo)$^b$   & 0.471 & 0.360 & 0.310 & 0.278 & 0.319 & 0.277 & 0.286 & 0.400 & 0.000 \\
		\bottomrule
	\end{tabular}
\label{tab:si_n1_top1_route_length}
  \raggedright
 \\\footnotesize{$^a$Steps not provided during inference.}
 \\\footnotesize{$^b$SMs not provided during inference.}
\end{table}

\begin{table}[htbp]
\caption{Top-10 Accuracy on n$_1$ routes split by route length}
	\centering
	\begin{tabular}{lccccccccc}
	\toprule
	\multicolumn{1}{c}{} & \multicolumn{9}{c}{Top10 accuracy on routes with length} \\
	Model            & 2     & 3     & 4     & 5     & 6     & 7     & 8     & 9     & 10    \\
	\midrule
        Explorer-XL$^{a, b}$  & 0.608 & 0.463 & 0.387 & 0.363 & 0.390 & 0.400 & 0.333 & 0.600 & 0.000 \\
	Deep$^b$         & 0.641 & 0.519 & 0.442 & 0.364 & 0.362 & 0.323 & 0.333 & 0.600 & 0.000 \\
	Explorer$^a$     & 0.692 & 0.546 & 0.417 & 0.373 & 0.376 & 0.431 & 0.333 & 0.500 & 0.000 \\
	Flash            & 0.763 & 0.600 & 0.465 & 0.385 & 0.408 & 0.400 & 0.333 & 0.500 & 0.000 \\
	Wide$^b$         & 0.647 & 0.531 & 0.434 & 0.378 & 0.390 & 0.354 & 0.333 & 0.600 & 0.000 \\
	Flex (Mono)      & 0.650 & 0.474 & 0.325 & 0.254 & 0.286 & 0.231 & 0.143 & 0.300 & 0.000 \\
	Flex (Duo)       & 0.765 & 0.610 & 0.469 & 0.373 & 0.408 & 0.385 & 0.333 & 0.600 & 0.000 \\
	Flex (Duo)$^b$   & 0.594 & 0.460 & 0.388 & 0.336 & 0.371 & 0.338 & 0.333 & 0.600 & 0.000 \\
	\bottomrule
\end{tabular}
\label{tab:si_n1_top10_route_length}
  \raggedright
 \\\footnotesize{$^a$Steps not provided during inference.}
 \\\footnotesize{$^b$SMs not provided during inference.}
\end{table}

\begin{table}[htbp]
\caption{Top-1 Accuracy on n$_5$ routes split by route length}
	\centering
	\begin{tabular}{lccccccccc}
	\toprule
	\multicolumn{1}{c}{} & \multicolumn{9}{c}{Top1 accuracy on routes with length} \\
	Model            & 2     & 3     & 4     & 5     & 6     & 7     & 8     & 9     & 10    \\
	\midrule
        Explorer-XL$^{a, b}$  & 0.395 & 0.317 & 0.284 & 0.303 & 0.313 & 0.248 & 0.405 & 0.692 & 0.333 \\
	Deep$^b$         & 0.474 & 0.384 & 0.343 & 0.313 & 0.338 & 0.272 & 0.381 & 0.385 & 0.333 \\
	Explorer$^a$     & 0.458 & 0.350 & 0.300 & 0.289 & 0.343 & 0.288 & 0.286 & 0.462 & 0.333 \\
	Flash            & 0.502 & 0.401 & 0.337 & 0.320 & 0.331 & 0.328 & 0.429 & 0.692 & 0.333 \\
	Wide$^b$         & 0.477 & 0.399 & 0.352 & 0.337 & 0.366 & 0.304 & 0.381 & 0.615 & 0.333 \\
	Flex (Mono)      & 0.436 & 0.324 & 0.258 & 0.211 & 0.191 & 0.184 & 0.190 & 0.154 & 0.333 \\
	Flex (Duo)       & 0.549 & 0.416 & 0.366 & 0.336 & 0.352 & 0.328 & 0.476 & 0.615 & 0.333 \\
	Flex (Duo)$^b$  & 0.465 & 0.358 & 0.324 & 0.312 & 0.326 & 0.344 & 0.429 & 0.615 & 0.333 \\
	\bottomrule
    \end{tabular}
\label{tab:si_n5_top1_route_length}
   \raggedright
 \\\footnotesize{$^a$Steps not provided during inference.}
 \\\footnotesize{$^b$SMs not provided during inference.}
\end{table}

\begin{table}[htbp]
\caption{Top-10 Accuracy on n$_5$ routes split by route length}
	\centering
	\begin{tabular}{lccccccccc}
	\toprule
	\multicolumn{1}{c}{} & \multicolumn{9}{c}{Top10 accuracy on routes with length} \\
	Model            & 2     & 3     & 4     & 5     & 6     & 7     & 8     & 9     & 10    \\
	\midrule
        Explorer-XL$^{a, b}$  & 0.586 & 0.469 & 0.416 & 0.404 & 0.423 & 0.384 & 0.47  & 0.692 & 0.333 \\
	Deep$^b$         & 0.604 & 0.507 & 0.451 & 0.414 & 0.430 & 0.368 & 0.429 & 0.615 & 0.333 \\
	Explorer$^a$     & 0.654 & 0.529 & 0.442 & 0.404 & 0.434 & 0.376 & 0.405 & 0.692 & 0.333 \\
	Flash            & 0.712 & 0.583 & 0.479 & 0.431 & 0.425 & 0.400 & 0.429 & 0.692 & 0.333 \\
	Wide$^b$         & 0.616 & 0.524 & 0.451 & 0.420 & 0.432 & 0.328 & 0.476 & 0.692 & 0.333 \\
	Flex (Mono)      & 0.621 & 0.461 & 0.357 & 0.285 & 0.297 & 0.272 & 0.333 & 0.462 & 0.333 \\
	Flex (Duo)       & 0.721 & 0.594 & 0.489 & 0.436 & 0.444 & 0.416 & 0.500 & 0.692 & 0.333 \\
	Flex (Duo)$^b$   & 0.581 & 0.462 & 0.407 & 0.383 & 0.421 & 0.400 & 0.452 & 0.692 & 0.333 \\
	\bottomrule
\end{tabular}
\label{tab:si_n5_top10_route_length}
   \raggedright
 \\\footnotesize{$^a$Steps not provided during inference.}
 \\\footnotesize{$^b$SMs not provided during inference.}
\end{table}

\begin{figure}[htbp]
  \centering
  \includegraphics[width=\textwidth]{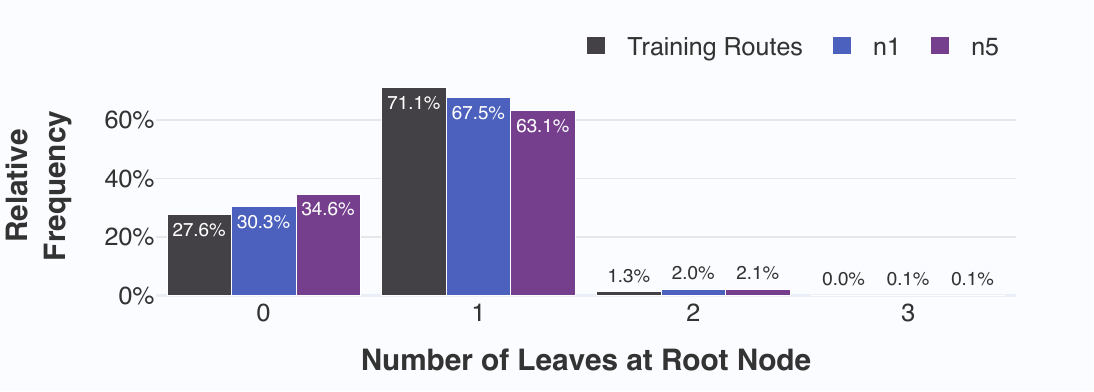}  
  \caption{Distribution of the relative frequencies for the number of leaves (nodes with no children) at root node.}
  \label{fig:leaf_dist}
\end{figure}

\begin{figure}[htbp]
  \centering
  \includegraphics[width=\textwidth]{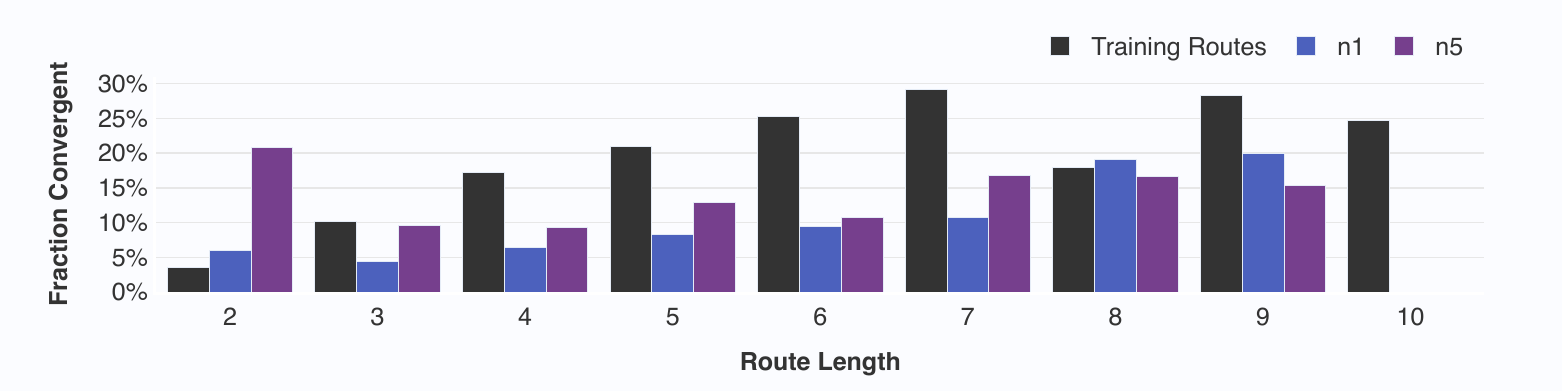}  
  \caption{Fraction of routes (grouped by route length) that are convergent in training set, n$_1$, and n$_5$. A route is considered convergent if any node has at least two children that are not leaves. Overall, convergent routes constitute 9.4\% of the training routes, 5.6\% of n$_1$, and 10.9\% of n$_5$.}
  \label{fig:conv_frac_by_length}
\end{figure}

\begin{figure}[htbp]
  \centering
  \includegraphics[width=\textwidth]{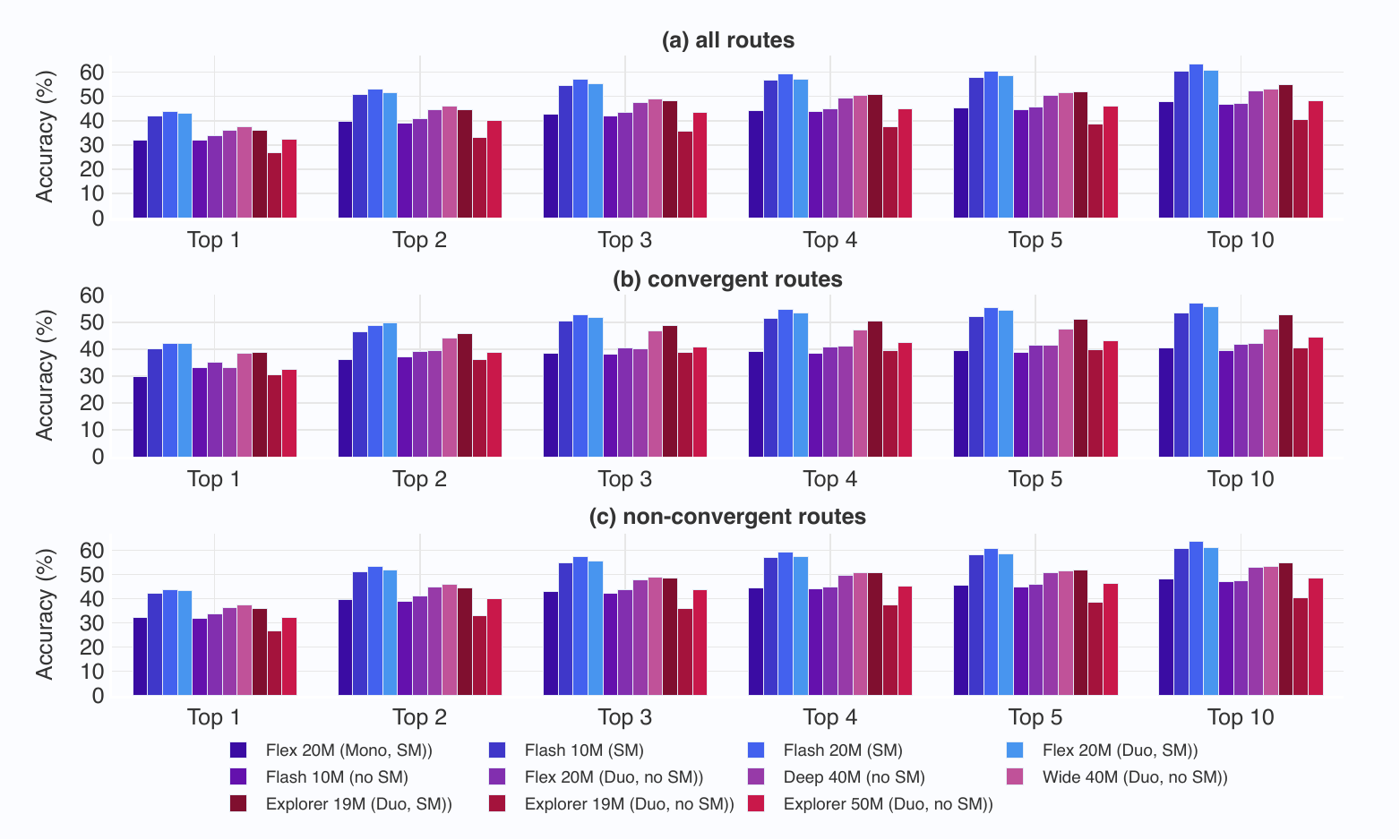}  
  \caption{Top-K accuracy on the n$_1$ test set for different models for (a) all routes in n$_1$, (b) convergent routes in n$_1$, and (c) non-convergent routes in n$_1$. A route is considered convergent if any node has at least two children that are not leaves.}
  \label{fig:n1_topk_subplots}
\end{figure}

\begin{figure}[htbp]
  \centering
  \includegraphics[width=\textwidth]{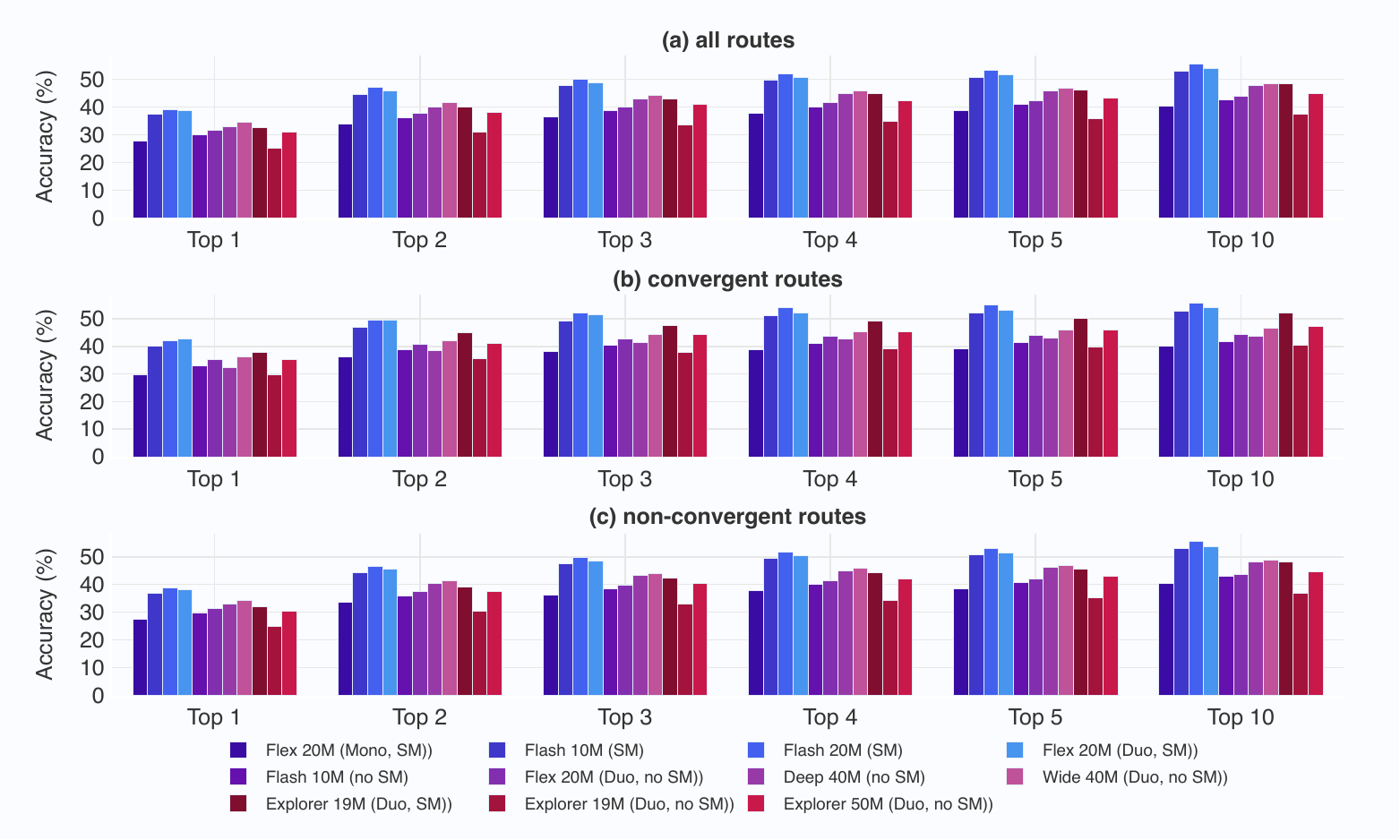}  
  \caption{Top-K accuracy on the n$_5$ test set for different models for (a) all routes in n$_5$, (b) convergent routes in n$_5$, and (c) non-convergent routes in n$_5$. A route is considered convergent if any node has at least two children that are not leaves.}
  \label{fig:n5_topk_subplots}
\end{figure}

\end{document}